\documentclass[a4paper,fleqn]{cas-sc}
\usepackage[authoryear,round,sort&compress]{natbib}
\usepackage{amssymb}
\usepackage{latexsym}
\usepackage{hyperref}

\usepackage{url}
\usepackage{xcolor}
\usepackage{multirow} 
\usepackage{pifont}
\usepackage{tabularx}
\usepackage{colortbl}
\usepackage{xcolor}
\usepackage{array}
\usepackage{algorithm}
\usepackage{algorithmic}
\usepackage{newfloat}
\usepackage{makecell}
\usepackage{listings}
\usepackage{float}
\usepackage{caption}
\usepackage{hyperref}
\usepackage{amsmath,amssymb,amsfonts}
\usepackage{booktabs}
\usepackage{amsmath}
\usepackage{amssymb}
\usepackage{enumitem}
\usepackage{xcolor}
\usepackage{setspace}
\usepackage{hyperref}
\usepackage{amsmath}
\usepackage{verbatim}

\def\tsc#1{\csdef{#1}{\textsc{\lowercase{#1}}\xspace}}
\tsc{WGM}
\tsc{QE}
\begin{document}
\let\WriteBookmarks\relax
\def\floatpagepagefraction{1}
\def\textpagefraction{.001}

\shorttitle{}    

\shortauthors{}  

\title [mode = title]{NeuroAlign: Hierarchical Multimodal Fusion of Dynamic and Structural Neuroimaging for MCI Analysis}  

\tnotemark[1] 

\tnotetext[1]{} 


% \author[1]{}%[<options>]
\author[1]{Xiongri Shen}
\author[2]{Zhenxi Song}
\cortext[cor2]{Corresponding author: 
songzhenxi@hit.edu.cn}
\author[1]{Jiaqi wang}
\author[1]{Yi Zhong}
\author[1]{Leilei Zhao}
 \author[3]{Chenqi Xu}
\author[4]{Linling Li }
\author[5]{Yichen Wei}
\author[5]{Lingyan Liang}
\author[5]{Demao Deng}
\author[6]{Luping Song}
\author[7]{Ping  Luan }
\author[8]{Ahmed M. Anter }
\author[10]{Shuqiang Wang}
\author[9]{Baiying Lei}
\author[2]{Zhiguo  Zhang}
\cortext[cor1]{Corresponding author: 
zhiguozhang@hit.edu.cn}
% \fntext[fn1]{This is author footnote for second author.}
% \author[2]{Given-name3 \snm{Surname3}}
% %% Third author's email
% \ead{author3@author.com}
% \author[2]{Given-name4 \snm{Surname4}}
\address[1]{Department of Computer Science and Technology, Harbin Institute of Technology (Shenzhen), Shenzhen, 518055, China}
\address[2]{School of Intelligence Science and Engineering, College of Artificial Intelligence, Harbin Institute of Technology, Shenzhen, 518055, China}
\address[3]{ School of Artificial Intelligence, Beijing University of Posts and Telecommunications, Beijing, China}
\address[4]{Guangdong Key Laboratory of Biomedical Measurements and Ultrasound Imaging, School of Biomedical Engineering, Shenzhen University Medical School, Shenzhen University, Shenzhen, 518055, China}
\address[5]{Department of Radiology, The People’s Hospital of Guangxi Zhuang Autonomous Region, Guangxi Academy of Medical Sciences, Guangxi Province, 518000, China}
\address[6]{Shenzhen Sxith People's Hospital (Nanshan Hospital), Huazhong University of Science and Technology Union Shenzhen Hospital, Shenzhen, 518055, China}
\address[7]{School of Basic Medical Sciences, Shenzhen University, Shenzhen, 518055, China}
\address[8]{the Egypt-Japan University of Science and Technology (E-JUST), Alexandria, 21934, Egypt}
\address[9]{School of Biomedical Engineering, National-Regional Key Technology Engineering Laboratory for Medical Ultrasound, Guangdong Key Laboratory for Biomedical, Measurements and Ultrasound Imaging, Shenzhen University Medical School, Shenzhen University, Shenzhen, 518055, China}
\address[10]{ the Shenzhen Institutes of Advanced Technology, Chinese Academy of Sciences, Shenzhen, 518055, China}

% \cormark[1]

% \ead{xiongrishen@stu.hit.edu.cn}


% \credit{
% \affiliation[1]{organization={},
% %          citysep={}, % Uncomment if no comma needed between city and postcode

% \author[2]{}%[]

% % Footnote of the second author

% % Email id of the second author
% \ead{}

% % URL of the second author


% \affiliation[2]{organization={},
% %          citysep={}, % Uncomment if no comma needed between city and postcode

% \cortext[1]{Corresponding author}

%Footnote text
\fntext[1]{Code and models are available at: \href{https://github.com/SXR3015/Interpretable-NeuroAlign}{NeuroAlign}}
\begin{abstract}
Multimodal neuroimaging fusion of functional MRI (fMRI) and diffusion tensor imaging (DTI) provides complementary information for cognitive impairment analysis, but remains challenged by heterogeneous feature spaces and misaligned representations. We propose \textit{NeuroAlign}, a hierarchical framework for structured multimodal fusion. It introduces (1) \textit{Dual-Modal Hierarchical Alignment} (DMHA), which models multi-scale dynamic connectivity and aligns dynamic--static and functional--structural embeddings; and (2) \textit{Dual-Domain Hierarchical Interaction} (DDHI), which enables fine-grained modulation and global interaction between connectivity- and region-level features. To support feature-level inspection, we design \textit{Synergistic Activation Mapping} (SAM), a gradient-free, marker-oriented attribution method for DFC, SFC, ALFF, and FA. Evaluated on GUTCM, ADNI, and OASIS under five-fold validation, NeuroAlign achieves competitive MCI/SCD detection and preliminary cross-dataset transferability. Attribution analyses reveal modality-specific and partially consistent brain patterns, providing model-derived evidence for multimodal representation analysis.

\end{abstract}

%    \begin{figure}
%     \centering
%     \includegraphics[width=0.99\textwidth]{Figure1.jpg}
%     \caption{
% Overview of \textbf{\textit{NeuroAlign}}: A hierarchical multimodal fusion framework for MCI analysis. \textit{DMHA} aligns dynamic (DFC, ALFF) and structural (SFC, FA) neuroimaging across temporal scales and modalities; \textit{DDHI} enables fine-grained and global interaction between regional and connectivity domains; \textit{SAM} estimates modality-specific attribution maps for DFC, SFC, ALFF, and FA.}

%     \label{fig: Fig1}
% \end{figure}


\begin{highlights}
    \item NeuroAlign achieves 0.825/0.735 accuracy on ADNI/OASIS in multimodal MCI classification.
    \item DFC+SFC+ALFF+FA fusion improves accuracy over the best single-feature setting.
    \item TSA captures dynamic functional connectivity patterns across multiple temporal scales.
    \item SAM estimates model-relevant DFC, SFC, ALFF, and FA marker attribution maps.
    \item Cross-site results show partial transferability with remaining domain gaps.
\end{highlights}

% Keywords
\begin{keywords}
 Functional-structural fusion\sep fMRI-DTI\sep Functional-structural connectivity \sep Cognitive impairment 
\end{keywords}

\maketitle

% Main text
\section{Introduction}

   \begin{figure}
    \centering
    \includegraphics[width=0.99\textwidth]{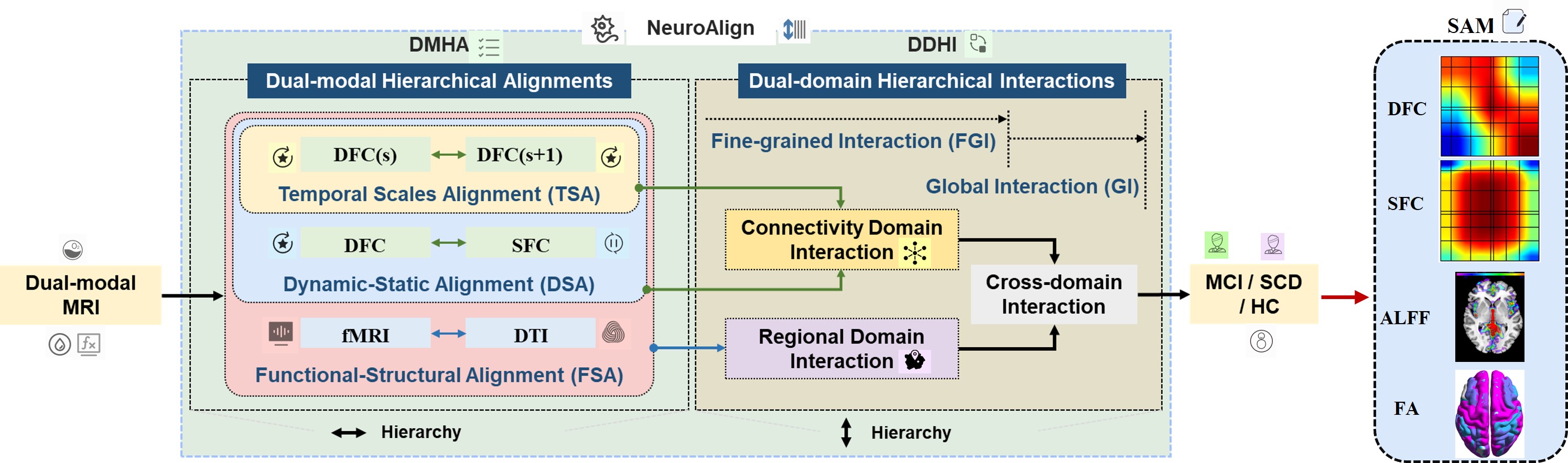}
    \caption{
    Overview of \textbf{\textit{NeuroAlign}}: A hierarchical multimodal fusion framework for MCI analysis. \textit{DMHA} aligns dynamic (DFC, ALFF) and structural (SFC, FA) neuroimaging across temporal scales and modalities; \textit{DDHI} enables fine-grained and global interaction between regional and connectivity domains; \textit{SAM} estimates modality-specific attribution maps for DFC, SFC, ALFF, and FA.}

    \label{fig: Fig1}
\end{figure}
Recent advances in neuroimaging have enabled non-invasive characterization of brain structure and function, offering rich data for multimodal fusion. Functional magnetic resonance imaging (fMRI) captures dynamic neural activity through blood-oxygen-level-dependent (BOLD) signals, while diffusion tensor imaging (DTI) reveals white matter integrity via fiber connectivity. These modalities provide complementary views of neurobiological changes associated with cognitive impairment; however, their effective integration remains an open challenge in information fusion due to misaligned representations, heterogeneous temporal scales, and domain-specific feature semantics.

\begin{figure}
    \centering
       \includegraphics[width=0.7\textwidth]{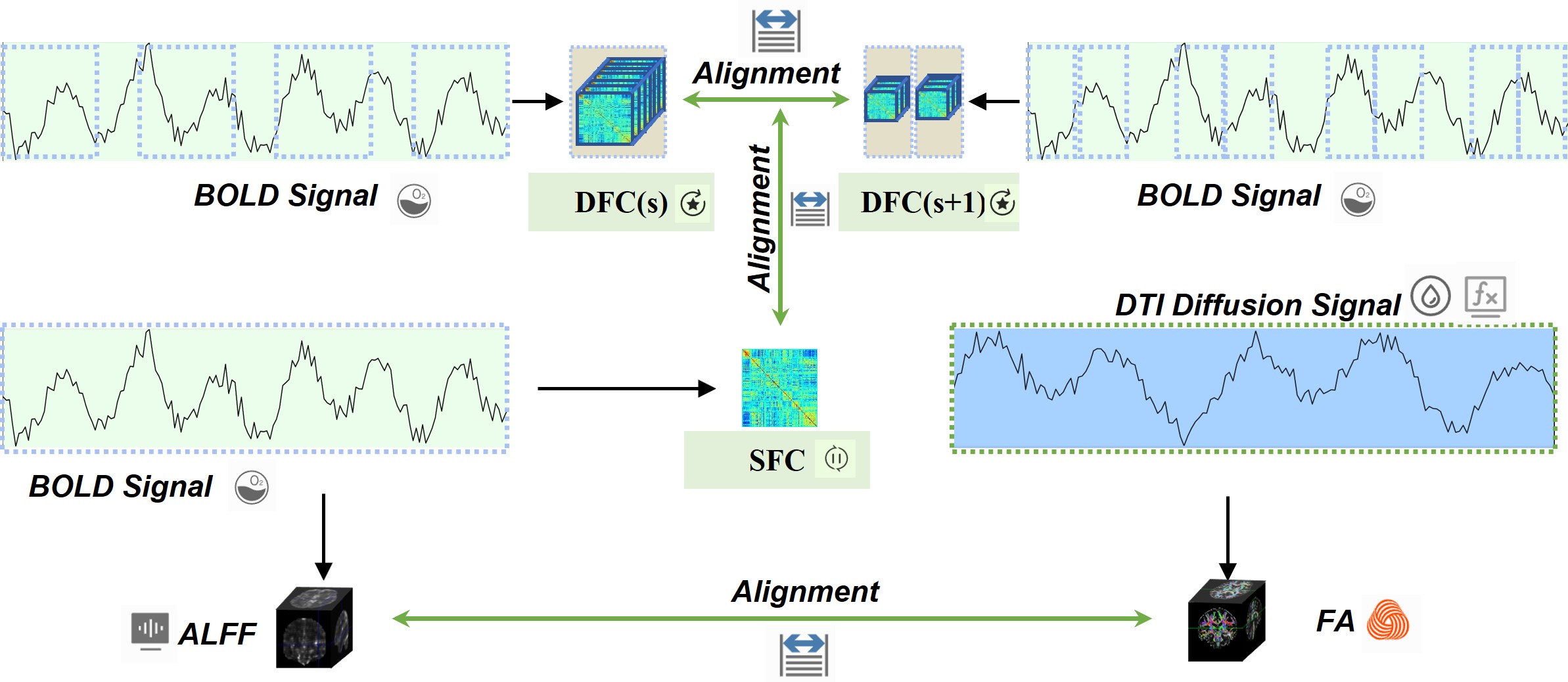}
    \caption{Overview of dual-modal hierarchical alignment in fMRI and DTI.}
    \label{fig: DMHA}
\end{figure}

\begin{figure}
    \centering
       \includegraphics[width=0.7\textwidth]{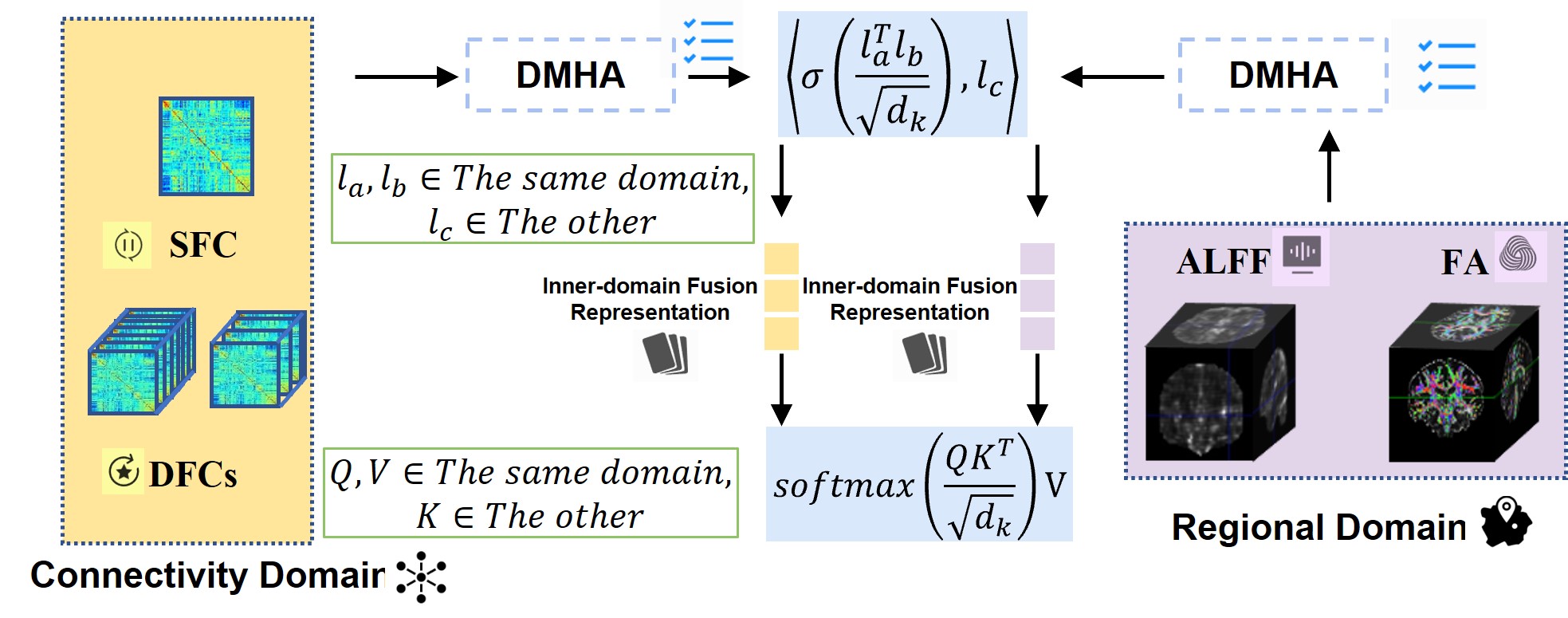}
    \caption{Overview of dual-domain hierarchical interaction in fMRI and DTI.}
    \label{fig: DDHI}
\end{figure}

A growing body of work leverages machine learning and deep learning to integrate fMRI- and DTI-derived features for automated analysis~\cite{cummings2020alzheimer,reuben2020memory,qiu2022multimodal,huang2025multiview,li2023alterations,liang2021recurrent}. Common approaches extract static functional connectivity (SFC), dynamic functional connectivity (DFC), amplitude of low-frequency fluctuations (ALFF), and fractional anisotropy (FA) as input representations. However, most existing frameworks treat these features independently—processing each modality separately and concatenating high-level embeddings without explicit cross-modal alignment. This limits their ability to model the complex, covariant nature of functional and structural degeneration in aging brains~\cite{zhang2023static,wang2022static}.

We identify four key limitations in current multimodal MRI analysis from an information fusion perspective:

\begin{enumerate}[label=(\arabic*)]
    \item \textbf{Lack of functional--structural alignment:} While fMRI and DTI provide complementary functional and structural information~\cite{xu2021different,avila2022brain,zhong2022shared}, their feature spaces differ substantially in scale, distribution, and semantics. Few methods explicitly align these representations, which may limit effective multimodal fusion.

    \item \textbf{Disjoint modeling of static and dynamic connectivity:} SFC represents relatively stable network organization, whereas DFC captures time-varying interactions~\cite{briend2020aberrant,kam2019deep}. Most studies analyze them separately, which may overlook complementary information between static and dynamic connectivity patterns.

    \item \textbf{Fixed temporal scales in DFC analysis:} Conventional DFC often relies on sliding windows with fixed durations, which may be insufficient for representing temporal patterns at different resolutions~\cite{duda2021validating,zhang2022test}. Evidence from high-temporal-resolution data suggests that connectivity changes can appear across multiple time scales~\cite{song2018biomarkers,shen2022self}, motivating scale-aware integration.

    \item \textbf{Limited feature-level attribution in fusion models:} Despite improved performance, many multimodal architectures provide limited information about which input features are emphasized by the trained model~\cite{castellano2024automated}. This motivates post-hoc attribution strategies that can separately inspect connectivity-level and region-level feature patterns.
\end{enumerate}

To address these challenges, we propose \textbf{\textit{NeuroAlign}}, a hierarchical framework for structured multimodal fusion of dynamic and structural neuroimaging with marker-oriented attribution. As illustrated in Fig.~\ref{fig: Fig1}--\ref{fig: DDHI}, \textit{NeuroAlign} consists of two core components:

(1)~\textit{Dual-Modal Hierarchical Alignment (DMHA)}: A pyramid-based architecture that aligns multi-scale DFCs across temporal resolutions (TSA), synchronizes dynamic and static functional patterns via contrastive learning (DSA), and harmonizes functional (fMRI) and structural (DTI) features through functional--structural alignment (FSA), encouraging more compatible multimodal representations.

(2)~\textit{Dual-Domain Hierarchical Interaction (DDHI)}: A two-stage attention mechanism that first performs fine-grained modulation (FI) within domains (e.g., region-guided connectivity), then enables global cross-domain integration (GI) via multi-head attention, supporting information exchange across connectivity and regional domains.

To support post-hoc feature inspection, we introduce \textit{Synergistic Activation Mapping} (SAM), a gradient-free, marker-oriented attribution strategy tailored to the four input representations used in this study. SAM estimates model-relevant attribution maps for DFC, SFC, ALFF, and FA, enabling qualitative assessment of feature usage across modalities.

The main contributions of this work are:
\begin{itemize}
    \item We present \textbf{\textit{NeuroAlign}}, a structured \textit{multimodal fusion} framework that models alignment and interaction across modalities (fMRI/DTI), temporal scales (multi-scale DFC), and domains (regional/connectivity), aiming to improve multimodal representation learning and diagnostic performance.
    \item We design a marker-oriented attribution module (\textit{SAM}) that estimates feature-level attribution maps for DFC, SFC, ALFF, and FA, supporting qualitative inspection of model-relevant connectivity and regional patterns.
    \item We validate our method on three heterogeneous datasets (GUTCM, ADNI, OASIS), showing competitive performance and preliminary cross-dataset transferability under varying acquisition protocols and scanner platforms.
\end{itemize}

\begin{figure}
    \centering
       \includegraphics[width=0.99\textwidth]{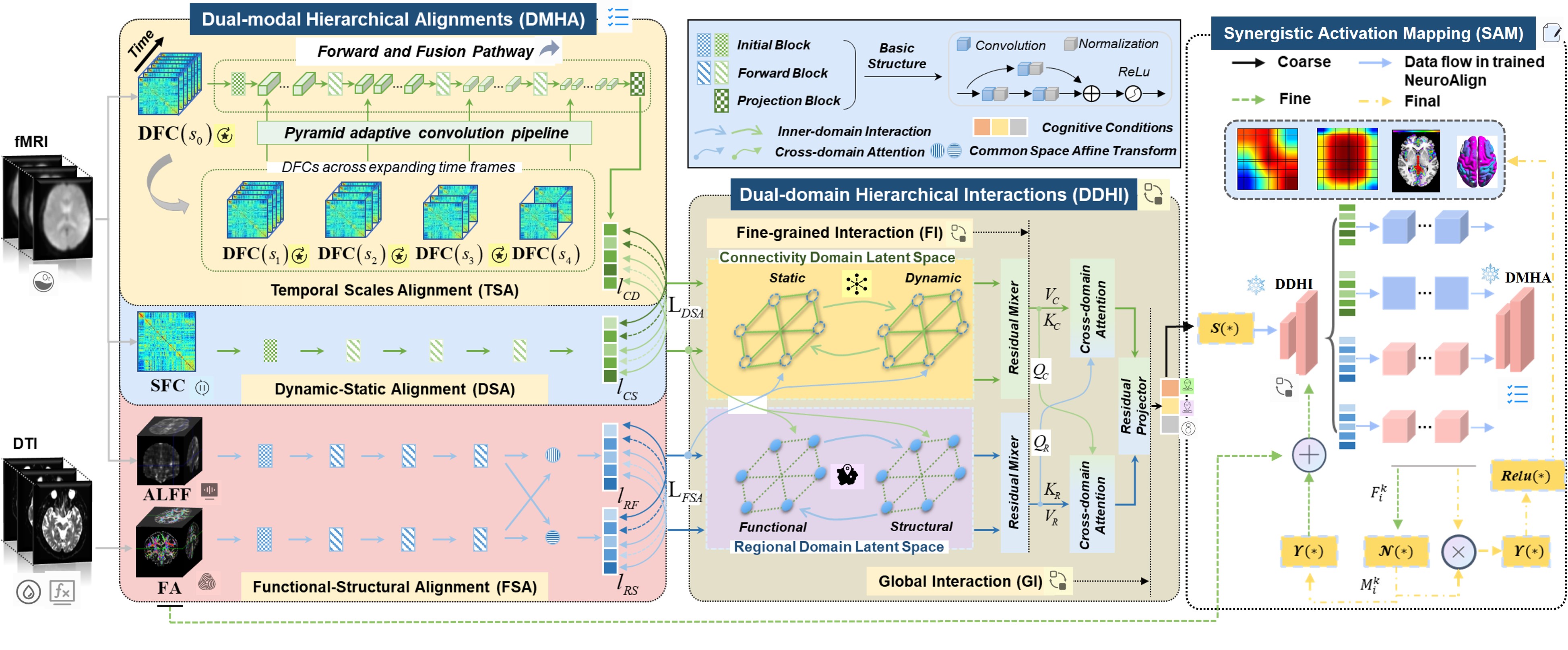}
    \caption{\textbf{\textit{NeuroAlign}} leverages fMRI and DTI inputs for cognitive impairment detection. It exploits \textit{DMHA} to perform hierarchical alignments across dynamic temporal scales, integrate dynamic and static networks, and correlate functional with structural features. \textit{DDHI} enables hierarchical feature fusion from fine-grained to global levels across regional and connectivity domains. \textit{SAM} estimates model-relevant DFC, SFC, ALFF, and FA attribution maps for feature-level inspection.}
    \label{fig: Fig2}
\end{figure}

\section{Related Works}

We categorize existing MRI-based approaches through the lens of \textit{information fusion} — from shallow integration to structured multimodal fusion with post-hoc attribution.

\textbf{Shallow fusion methods.} Early works rely on handcrafted features and classical classifiers, or perform late-stage concatenation of modality-specific embeddings~\cite{jeon2020enriched,dong2021integration,cao2023novel,wang2020multi,chen2023orthogonal}. While computationally efficient, these methods suffer from \textit{cross-modal misalignment}: fMRI-derived (e.g., ALFF, SFC) and DTI-derived (e.g., FA) features reside in heterogeneous spaces, and simple concatenation fails to model their covariant relationships~\cite{chen2023orthogonal,wang2020multi}. This limits representation discriminability and generalization across sites.

\textbf{Deep fusion methods.} Recent deep learning frameworks adopt end-to-end pipelines that jointly process fMRI and DTI inputs~\cite{gao2023hybrid,he2024spatiotemporal,faghiri2020weighted,li2025transformer}. These include CNNs for regional feature extraction, RNNs/LSTMs for temporal modeling of DFC, and transformers for long-range connectivity capture. However, most treat modalities as independent channels, neglecting explicit alignment between dynamic (fMRI) and structural (DTI) representations — leading to suboptimal synergy exploitation~\cite{zhang2023static,wang2022static}. Further, they often assume fixed temporal scales (e.g., 30s DFC windows), ignoring multi-scale dynamics critical to early impairment detection~\cite{duda2021validating,zhang2022test}.

\textbf{Structured and interpretable fusion methods.} A growing body of work seeks principled mechanisms for cross-modal interaction and transparency. Some introduce contrastive objectives to align functional and structural embeddings~\cite{xu2021different,zhong2022shared}, while others design attention modules to weight inter-modality contributions~\cite{qiu2022multimodal,huang2025multiview}. Yet, these remain largely \textit{domain-agnostic}: they do not distinguish between regional activity (e.g., ALFF, FA) and connectivity patterns (e.g., SFC, DFC), nor do they explicitly model hierarchical coupling across them. Moreover, feature attribution remains challenging: Grad-CAM and its variants are often applied post-hoc to single-modal networks~\cite{selvaraju2017grad,castellano2024automated,huang2025pcg}, while heterogeneous inputs such as connectivity matrices and 3D regional maps require modality-specific visualization and careful interpretation.

In summary, current methods fall short in three key aspects of multimodal fusion: (i) \textit{explicit alignment} across modalities (fMRI/DTI), temporal scales (multi-scale DFC), and domains (regional/connectivity); (ii) \textit{structured interaction} between connectivity-level and region-level features; and (iii) \textit{feature-level attribution} for inspecting model-relevant functional--structural patterns. NeuroAlign is designed to address these aspects.

% \textbf{Interpretable methods in MRI}. 
% There are few model-driven explainable methods due to the high structural and high-dimensional characteristics of MRI. Recent research has directly applied Grad-CAM to MRI, but these studies lack robust designs for dual-modal architectures \cite{castellano2024automated,huang2025pcg}. Grad-CAM generates activation maps by directly backpropagating classification labels \cite{selvaraju2017grad}. A more robust and effective design is particularly needed for the high-structural dual-modal features.

\section{Proposed Method}
\label{sec:method}

We present \textbf{\textit{NeuroAlign}}, a hierarchical framework for \textit{structured multimodal fusion} of dynamic and structural neuroimaging to support cognitive impairment analysis. As illustrated in Fig.~\ref{fig: Fig2}, \textit{NeuroAlign} comprises three synergistic components:
\begin{enumerate}[leftmargin=*,noitemsep]
    \item \textbf{Dual-Modal Hierarchical Alignment (DMHA)}: Aligns multi-scale dynamics and static topology across temporal resolutions and modalities;
    \item \textbf{Dual-Domain Hierarchical Interaction (DDHI)}: Enables fine-grained and global cross-domain communication between connectivity and regional features;
    \item \textbf{Synergistic Activation Mapping (SAM)}: A marker-oriented attribution module that estimates model-relevant DFC, SFC, ALFF, and FA attribution maps.
\end{enumerate}
This design addresses three core challenges in multimodal neuroimaging fusion:
\begin{itemize}[leftmargin=*,noitemsep]
    \item \textbf{Time-scale heterogeneity}: Brain connectivity evolves at multiple temporal scales — fixed-length windows (e.g., 30s) miss transient, disease-relevant motifs like epileptiform bursts or compensatory reorganization.
    \item \textbf{Functional-static complementarity}: Dynamic (DFC) and static (SFC) networks encode distinct but aligned cognitive states — DFC reveals ``how'' cognition unfolds; SFC reveals ``where'' it resides anatomically.
    \item \textbf{Cross-modal complementarity}: Functional (fMRI) and structural (DTI) data provide complementary information, with fMRI reflecting activity-related signals and DTI reflecting diffusion-related structural properties.
\end{itemize}

\subsection{Dual-Modal Hierarchical Alignment (DMHA)}
\label{sec:dmha}

DMHA supports \textit{structured multimodal fusion} by harmonizing representations across time scales, modalities, and domains, reducing reliance on a single manually selected temporal window.

\subsubsection{TSA: Multi-Scale Temporal Alignment via Pyramid Pooling}
\label{sec:tsa}

\textbf{Fusion challenge:} Fixed sliding windows fail to capture both transient events (e.g., 5s hyperconnectivity spikes) and long-term trends (e.g., 90s slow-wave coupling), limiting dynamic pattern modeling.  
\textbf{Fusion solution:} We construct a pyramid of DFCs using sliding windows from 20s to 100s (step = 20s), generating 5 scale-specific tensors $\mathcal{D}_{s_l} \in \mathbb{R}^{W \times H \times T_{s_l}}$. Each tensor is encoded by a spatial-strided CNN, downsampling feature maps by $2^{-l}$ in space and $4^{-l}$ in time.

Crucially, we avoid heuristic fusion. Instead, each $\mathcal{D}_{s_l}$ is projected into a shared latent space via a lightweight adapter $f_s(\cdot)$, then concatenated along the channel dimension:
\begin{equation}
    F_{\text{cat}} = \mathrm{Concat}_l \left[ f_s(\mathcal{D}_{s_l}) \right], \quad 
    \text{where } f_s(\cdot) \text{ is a 1×1 conv + GELU}.
\end{equation}
A depth-wise refinement module $g_s(\cdot)$ synchronizes the fused representation with the forward pathway:
\begin{equation}
    l_{CD} = g_s(F_{\text{cat}}), \quad 
    \text{where } g_s(\cdot) \text{ is a depth-wise separable convolution}.
\end{equation}
\textbf{Fusion outcome:} $l_{CD}$ encodes multi-scale temporal dynamics as a single, unified connectivity embedding — enabling joint modeling of short-term fluctuations and long-term trends within a single fusion pipeline.

\begin{figure}
    \centering 
    \includegraphics[width=0.99\textwidth]{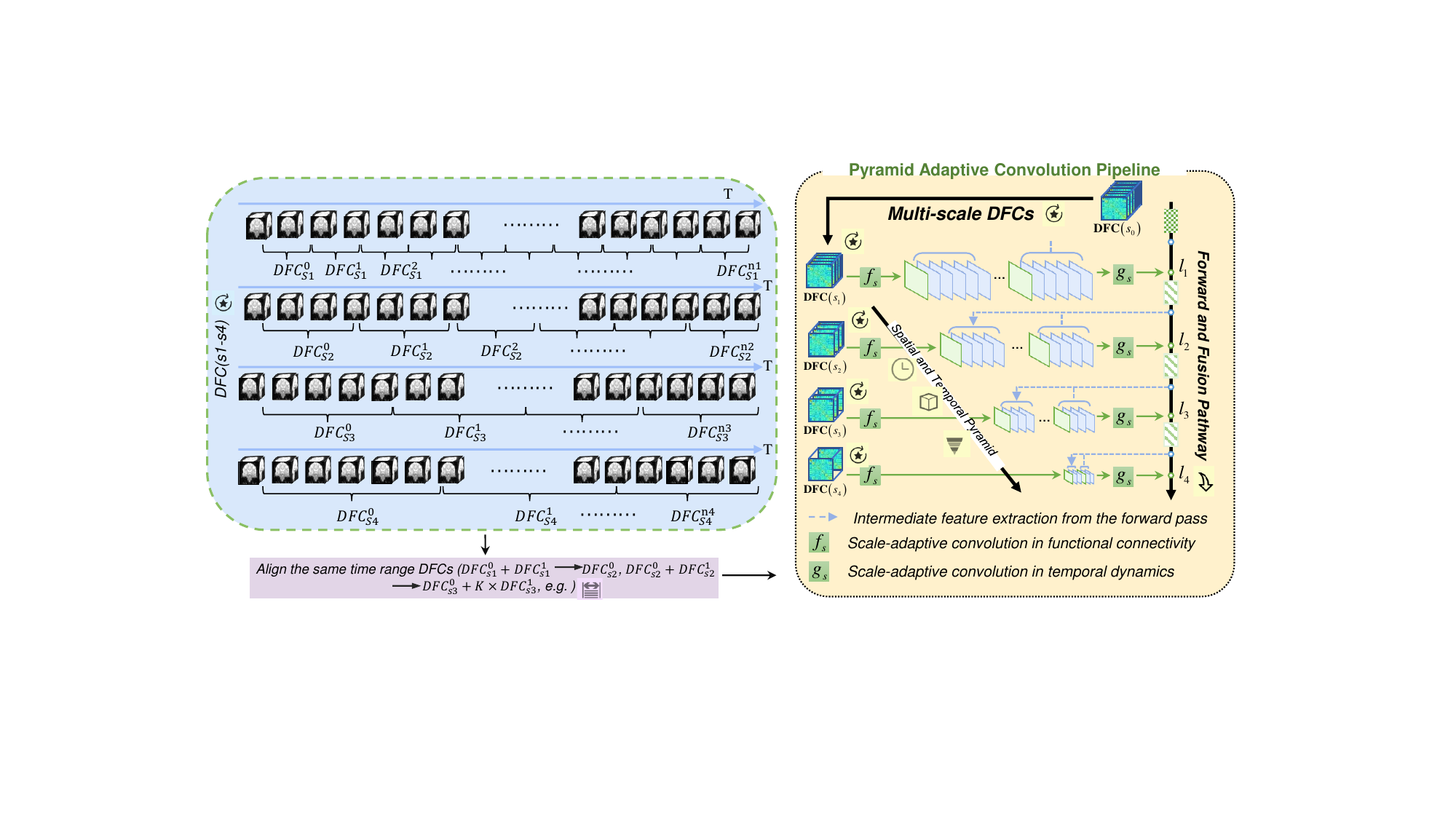}
    \caption{Multi-scale DFC alignment in TSA. Sliding windows (20s–100s) generate scale-specific DFCs, which are adaptively fused into a unified connectivity embedding $l_{CD}$.}
    \label{fig: Fig3}
\end{figure} 

\begin{algorithm}[t]
\small
\caption{Multi-scale DFC alignment in TSA}
\label{alg:tsa}
\begin{algorithmic}[1]
\REQUIRE DFC tensors $\{ \mathcal{D}_{s_0}, \dots, \mathcal{D}_{s_4} \}$ (20s to 100s); adapter $f_s$; refinement $g_s$; strided convolutions $C_{2d}, C_0$
\ENSURE Unified connectivity embedding $l_{CD}$
\STATE $M_0 = C_0(\mathcal{D}_{s_0})$, $D_0 = \mathrm{depth}(M_0)$ \COMMENT{Base-level feature map}
\FOR{$i = 1$ to $4$}
    \STATE $M_i = f_s(\mathcal{D}_{s_i})$, $D_i = \mathrm{depth}(M_i)$ \COMMENT{Project scale-$i$ DFC to shared space}
    \STATE $K_i = \mathrm{round}(D_{i-1}/D_i)$ \COMMENT{Depth ratio for channel alignment}
    \FOR{$d_i = 1$ to $D_i$}
        \FOR{$z_i = (d_i{-}1)K_i$ to $d_i K_i$}
            \STATE $M_{z_i} = C_{2d}[M_i(d_i), M_0(z_i)]$ \COMMENT{Cross-scale channel fusion}
        \ENDFOR
    \ENDFOR
    \STATE $F_i = g_s(\mathrm{Concat}(M_{z_i}))$ \COMMENT{Refine fused features}
\ENDFOR
\STATE \textbf{return} $F_4$ \COMMENT{Final multi-scale embedding $l_{CD}$}
\end{algorithmic}
\end{algorithm}

\subsubsection{DSA: Dynamic-Static Contrastive Alignment}
\label{sec:dsa}

\textbf{Fusion challenge:} DFC and SFC originate from different acquisition protocols and preprocessing pipelines — their embeddings lie in misaligned semantic spaces. Naive concatenation fails to capture subject-level consistency, harming fusion discriminability.  
\textbf{Fusion solution:} We extract high-level embeddings $z_d = \phi_d(x_{\text{DFC}})$ and $z_s = \phi_s(x_{\text{SFC}})$, then apply contrastive learning to pull same-subject pairs closer while pushing cross-subject pairs apart:
\begin{equation}
    \mathcal{L}_{\text{DSA}} = -\log \frac{\exp(\mathrm{sim}(z_d, z_s)/\tau)}{\sum_{j=1}^{2N} \mathbf{1}_{[j \neq d]} \exp(\mathrm{sim}(z_d, z_j)/\tau)},
\end{equation}
where $\tau = 0.07$ is a learnable temperature scalar and $\mathrm{sim}(a,b) = \frac{a^\top b}{\|a\|\|b\|}$.  
\textbf{Fusion outcome:} This enforces \textit{subject-level alignment} — ensuring that for Patient A, $z_d^A$ and $z_s^A$ are semantically closer than $z_d^A$ and $z_s^B$ (Patient B), preserving subject-level connectivity information that may be useful for multimodal fusion.

\subsubsection{FSA: Functional-Structural Fusion via Affine-Invariant Learning}
\label{sec:fsa}

\textbf{Fusion challenge:} Standard fusion operations such as concatenation, summation, or gating may bias the model toward a specific arithmetic interaction and may not sufficiently account for functional--structural heterogeneity across cohorts.  
\textbf{Fusion solution:} We treat fusion as an \textit{invariance learning problem}. Given functional $z_F$ and structural $z_S$ features, we form two complementary probes:
\begin{equation}
    Z_* = z_F \odot z_S \quad (\text{element-wise product}), \quad 
    Z_+ = z_F + z_S \quad (\text{addition}),
\end{equation}
and encourage the model to align these complementary probes from the same subject:
\begin{equation}
    \mathcal{L}_{\text{FSA}} = -\log \frac{\exp(\mathrm{sim}(Z_*, Z_+)/\tau)}{\sum_{i=1}^{2N} \mathbf{1}_{[i \neq *]} \exp(\mathrm{sim}(Z_*, Z_i)/\tau)}.
\end{equation}
\textbf{Fusion outcome:} The model is encouraged to reduce sensitivity to a specific arithmetic fusion rule and to learn more compatible functional--structural representations. This design may help improve stability under heterogeneous acquisition settings, although it does not eliminate domain shifts.

\subsection{Dual-Domain Hierarchical Interaction (DDHI)}
\label{sec:ddhi}

DDHI bridges two complementary domains: \textit{connectivity} ($l_{CD}, l_{CS}$) and \textit{regional} ($l_{RF}, l_{RS}$), where subscripts $D$/S denote dynamic/static, $F$/S denote functional/structural — enabling structured cross-domain fusion.

\subsubsection{FI: Fine-grained Modulated Attention}
\label{sec:fi}

\textbf{Fusion challenge:} Region-level features and connectivity-level features provide complementary information, but static attention may not explicitly condition one feature domain on another.  
\textbf{Fusion solution:} We introduce \textit{modulated attention}: one domain’s embedding acts as a top-down controller for another’s intra-domain interaction. For example:
\begin{equation}
    I_{\text{FG}}(l_{RF}, l_{CS}, l_{RF}) = \left\langle \sigma\left( \frac{l_{CS}^\top l_{RF}}{\sqrt{d_k}} \right), l_{RF} \right\rangle,
\end{equation}
where $l_{CS}$ (static connectivity) modulates how $l_{RF}$ (functional regional) attends to itself.  
\textbf{Fusion outcome:} The modulated attention mechanism provides feature-conditioned routing between regional and connectivity representations, allowing the model to evaluate region-level information in the context of connectivity-level features.

\subsubsection{GI: Global Cross-domain Attention}
\label{sec:gi}

\textbf{Fusion challenge:} Pre-defined pathways (e.g., “connectivity $\rightarrow$ regional”) may restrict potential interactions between heterogeneous feature domains.  
\textbf{Fusion solution:} We treat all four latents $\{l_{CD}, l_{CS}, l_{RF}, l_{RS}\}$ as tokens in a sequence. Using standard multi-head attention:
\begin{equation}
    I_{\text{G}} = \mathrm{Softmax}\left(\frac{Q K^\top}{\sqrt{d_k}}\right)V,
    \quad Q,K,V = \mathrm{MLP}(l_*),
\end{equation}
we allow all-to-all cross-domain communication.  
\textbf{Fusion outcome:} This design enables data-driven information exchange among the four latent representations and allows the model to capture task-relevant cross-domain relationships without imposing a fixed interaction order.

\subsection{Synergistic Activation Mapping for Single- and Multi-modal Biomarker Attribution}
\label{sec:sam}

To support post-hoc interpretation of the trained multimodal model, we introduce \textit{Synergistic Activation Mapping} (SAM), a gradient-free attribution strategy for estimating model-relevant candidate marker patterns from four neuroimaging representations: DFC, SFC, ALFF, and FA. Different from directly claiming clinical causality, SAM is designed to identify candidate attribution patterns that are emphasized by the trained model during classification. These patterns are interpreted as model-derived attributions rather than validated clinical or causal biomarkers.

SAM is applied at two complementary levels. First, for single-modal interpretation, SAM estimates feature-specific attribution maps for each individual representation. For DFC and SFC, the attribution maps highlight ROI-to-ROI connectivity patterns or network-level connections that contribute to the model output. For ALFF and FA, the attribution maps highlight regional functional activity and structural diffusion patterns, respectively. Second, for multi-modal interpretation, SAM compares the attribution maps across DFC, SFC, ALFF, and FA to examine whether different modalities emphasize convergent or complementary brain regions and networks.

Given an input subject with four feature types
$\mathcal{X}=\{X_{\mathrm{DFC}}, X_{\mathrm{SFC}}, X_{\mathrm{ALFF}}, X_{\mathrm{FA}}\}$,
SAM extracts an intermediate feature map $F_m$ for each modality $m \in \{\mathrm{DFC}, \mathrm{SFC}, \mathrm{ALFF}, \mathrm{FA}\}$. Channel-wise masks are generated by normalizing each channel:
\begin{equation}
    M_{m,k} = \mathcal{N}(F_{m,k}),
\end{equation}
where $k$ denotes the channel index and $\mathcal{N}(\cdot)$ denotes min--max normalization.

Each mask is resized to the input feature space through a modality-specific operator $\Upsilon_m(\cdot)$. For DFC and SFC, the resized mask is projected to the ROI-by-ROI connectivity space. For ALFF and FA, the resized mask is projected to the corresponding 3D brain volume. The masked input is then obtained as
\begin{equation}
    \widetilde{X}_{m,k} = X_m \odot \Upsilon_m(M_{m,k}),
\end{equation}
where $\odot$ denotes element-wise multiplication.

The perturbed multimodal input is forwarded through the trained model while the other modalities remain unchanged, and the class-specific response is recorded:
\begin{equation}
    s_{m,k}^{c}=P(y=c \mid X_{\mathrm{DFC}},X_{\mathrm{SFC}},X_{\mathrm{ALFF}},X_{\mathrm{FA}};\widetilde{X}_{m,k}).
\end{equation}
The final attribution map for modality $m$ and class $c$ is computed as
\begin{equation}
    A_m^c = \mathrm{ReLU}\left(\sum_k \alpha_{m,k}^{c} \cdot \Upsilon_m(M_{m,k})\right),
\end{equation}
where $\alpha_{m,k}^{c}$ is obtained by normalizing the class scores $\{s_{m,k}^{c}\}_{k}$ across channels.

The output of SAM is a set of modality-specific attribution maps:
\begin{equation}
    \mathcal{A}^{c}=
    \{A_{\mathrm{DFC}}^{c}, A_{\mathrm{SFC}}^{c}, A_{\mathrm{ALFF}}^{c}, A_{\mathrm{FA}}^{c}\}.
\end{equation}

In this study, SAM is used as a marker-oriented visualization tool for qualitative interpretation. Its goal is to inspect how different neuroimaging features are emphasized by the trained model, rather than to prove clinical utility or causal pathological mechanisms. We also note that SAM shares the general gradient-free re-inference idea with methods such as Score-CAM, but differs in its application target: SAM is adapted to heterogeneous neuroimaging representations, including connectivity matrices and 3D regional maps, and is used to compare both single-modal and multi-modal attribution patterns.

\section{Experimental}
\label{sec:5}

% \subsection{ Description of GUTCM, ADNI and OASIS}
% \label{sec: description}

% To validate the generalizability of \textbf{\textit{NeuroAlign}}, we utilized resting-state fMRI and DTI data from three distinct sources: \textbf{(1)} Data collected from the hospital, the First Affiliated Hospital of Guangxi University of Traditional Chinese Medicine, hereafter referred to as the 'GUTCM Dataset', \textbf{(2)} The Alzheimer's Disease Neuroimaging Initiative (ADNI) repository \cite{petersen2010alzheimer}, (3) Open Access Series of Imaging Studies (OASIS).

% To demonstrate NeuroAlign's generalizability, it's crucial to consider significant variations in participant demographics and MRI acquisition parameters across the two datasets. The differences include variations in participant ages, weights,  Montreal Cognitive Assessment (MOCA), and Minimum Mental State Examination (MMSE) scores, as well as disparities in MRI acquisition parameters like flip angle, percent phrase field of view, layer numbers, and thickness. The detailed demographics and MRI parameters of GUTCM, ADNI and OASIS are shown in Table \ref{tab:infor}.
\subsection{Datasets and Participant Characteristics}
\label{sec:datasets}

To evaluate the performance and preliminary transferability of \textbf{\textit{NeuroAlign}} across diverse populations and acquisition protocols, we conducted experiments on three independent multimodal neuroimaging datasets: 
(1)~a locally acquired cohort from the First Affiliated Hospital of Guangxi University of Traditional Chinese Medicine (\textbf{GUTCM}), 
(2)~the Alzheimer's Disease Neuroimaging Initiative dataset (\textbf{ADNI})~\cite{petersen2010alzheimer}, and 
(3)~the Open Access Series of Imaging Studies (\textbf{OASIS}).

The GUTCM dataset includes resting-state fMRI and diffusion tensor imaging (DTI) data from 251 participants. All scans were performed on a 3.0T Siemens Prisma scanner with informed consent approved by the local ethics committee. 
ADNI and OASIS are publicly available datasets widely used in brain aging and dementia research. ADNI contains high-quality multimodal MRI from early Alzheimer’s disease, late MCI, and healthy controls; OASIS provides longitudinal structural and functional imaging with detailed clinical assessments. After applying the paired multimodal inclusion criteria, the OASIS subset used in this study included 90 subjects, consisting of 45 NC and 45 MCI participants.

Demographic and scanning parameters—including age, gender, MMSE, MoCA scores, field strength, TR/TE, flip angle, slice thickness, and number of volumes—are summarized in Table~\ref{tab:infor}. 
Significant heterogeneity exists across sites in both subject characteristics and acquisition settings, making this a challenging setting for preliminary cross-site evaluation. Although the expanded OASIS subset provides a more stable external evaluation setting, its scale remains smaller than GUTCM and ADNI.

\begin{table*}[htbp]
\caption{Demographic and MRI acquisition parameters for the GUTCM, ADNI, and OASIS datasets. Values are presented as mean $\pm$ standard deviation unless otherwise specified.}
\label{tab:infor}
\centering
\footnotesize
\setlength{\tabcolsep}{4pt}
\renewcommand{\arraystretch}{1.25}

\begin{tabularx}{\textwidth}{
>{\centering\arraybackslash}p{1.15cm}
>{\centering\arraybackslash}p{0.75cm}
>{\centering\arraybackslash}p{1.15cm}
>{\centering\arraybackslash}X
>{\centering\arraybackslash}X
>{\centering\arraybackslash}X
>{\centering\arraybackslash}X
>{\centering\arraybackslash}X}
\toprule
\textbf{Dataset} & \textbf{Group} & \textbf{N (F/M)} & \textbf{Age (years)} & \textbf{MMSE} & \textbf{MoCA} & \textbf{Education (years)} & \textbf{Weight (kg)} \\
\midrule
GUTCM & NC   & 44/33 & $64.48 \pm 5.73$ & $29.55 \pm 0.72$ & $26.50 \pm 2.10$ & $12.15 \pm 2.94$ & $85.30 \pm 5.34$ \\
      & SCD  & 45/30 & $65.24 \pm 5.56$ & $27.00 \pm 0.87$ & $23.20 \pm 2.55$ & $11.04 \pm 2.80$ & $85.60 \pm 5.41$ \\
      & MCI  & 70/29 & $65.31 \pm 6.70$ & $25.31 \pm 1.04$ & $21.06 \pm 2.75$ & $10.45 \pm 2.94$ & $84.28 \pm 5.02$ \\
\addlinespace[5pt]
ADNI  & NC   & 38/28 & $76.70 \pm 6.47$ & $29.00 \pm 1.41$ & -- & -- & $79.29 \pm 13.09$ \\
      & MCI  & 41/55 & $77.40 \pm 7.07$ & $27.56 \pm 1.89$ & -- & -- & $79.10 \pm 13.99$ \\
\addlinespace[5pt]
OASIS & NC   & 25/20 & $69.65 \pm 4.27$ & $29.35 \pm 1.19$ & -- & $12.53 \pm 3.56$ & -- \\
      & MCI  & 25/20 & $74.50 \pm 8.62$ & $26.83 \pm 1.52$ & -- & $12.33 \pm 3.05$ & -- \\
\bottomrule
\end{tabularx}

\vspace{6mm}

\begin{tabularx}{\textwidth}{
>{\centering\arraybackslash}p{1.15cm}
>{\centering\arraybackslash}p{1.05cm}
>{\centering\arraybackslash}X
>{\centering\arraybackslash}X
>{\centering\arraybackslash}X
>{\centering\arraybackslash}X
>{\centering\arraybackslash}X
>{\centering\arraybackslash}X}
\toprule
\textbf{Dataset} & \textbf{Modality} & \textbf{TR (ms)} & \textbf{TE (ms)} & \textbf{FOV (mm)} & \textbf{Slice Thickness (mm)} & \textbf{Flip Angle} & \textbf{Slices} \\
\midrule
GUTCM & fMRI & 2000  & 230 & $100 \times 100$ & 5.0 & 90$^\circ$ & 31 \\
      & DTI  & 6800  & 93  & $256 \times 256$ & 3.0 & 90$^\circ$ & 46 \\
\addlinespace[5pt]
ADNI  & fMRI & 2000  & 30  & $93 \times 93$   & 3.3 & 80$^\circ$ & 31/48 \\
      & DTI  & 6800  & 30   & $93 \times 93$   & 1.2 & 9$^\circ$  & 16 \\
\addlinespace[5pt]
OASIS & fMRI & 2000  & 27  & $100 \times 100$ & 4.0 & 90$^\circ$ & 36 \\
      & DTI  & 14500 & 112 & $100 \times 100$ & 2.0 & 90$^\circ$ & 80 \\
\bottomrule
\end{tabularx}
\end{table*}
% \subsection{Data Preprocessing}
% \label{sec:preproccessing}
% The raw fMRI data underwent preprocessing routine using the SPM12 toolbox \cite{tzourio2002automated}. These steps included: 1) slice-timing correction, 2) head motion estimation and correction, 3) intra-subject registration, 4) co-registration, and 5) regression-based outlier removal. In the domain of connectivity analysis, the Dosenbach164 atlas was employed to extract Regions of Interest (ROIs) for reconstructing functional connectivity, encompassing both SFC and DFC patterns. The correlation between ROIs of Dosenbach164 is computed using Pearson's correlation. Given the temporal resolution of MRI scanner, we obtained a total of 160 dynamic temporal points in DFCs, with each point representing a functional connectivity frame.  Additionally, the ALFF between 0.01 $\sim$\ 0.1 Hz was extracted using the DPARSF tool \cite{yan2010dparsf} to illustrate regional characteristics. 

% For DTI preprocessing, the steps involve brain extraction using Bet, correction for eddy currents, probabilistic tractography using Probtrackx, and fitting the diffusion tensor model using Dtifit. These steps are carried out by FSL. We used the PANDA toolbox \cite{cui2013panda} to extract fiber tracts, where the Anatomical Automatic Labeling template was employed to guide ROI segmentation. The fiber strength between two ROIs was estimated based on the number of fibers. This was then normalized by the average surface area between the white and gray matter of those brain regions to derive the input features.

\subsection{Data Preprocessing Pipeline}
\label{sec:preprocessing}

% All raw imaging data were preprocessed using standardized pipelines to extract comparable features while preserving site-specific characteristics.

\subsubsection{Functional MRI Processing}

Resting-state fMRI data were processed using SPM12 (\url{https://www.fil.ion.ucl.ac.uk/spm/}) within MATLAB R2021b. The pipeline included: 
(i) slice-timing correction; 
(ii) rigid-body motion correction (threshold: $< $2 mm translation,  $< $ 2$^\circ$ rotation); 
(iii) co-registration of EPI to T1-weighted anatomical image; 
(iv) normalization to MNI space via DARTEL; 
(v) spatial smoothing with an 8-mm FWHM Gaussian kernel; 
(vi) band-pass temporal filtering (0.01–0.1 Hz) to retain low-frequency fluctuations.

For functional connectivity analysis, we adopted the Dosenbach-164 atlas, which parcellates the cortex into 164 regions covering six canonical networks (default mode, frontoparietal, cingulo-opercular, etc.). Time series were extracted for each ROI, and static functional connectivity (SFC) was computed as the Pearson correlation matrix across all region pairs. Dynamic functional connectivity (DFC) was estimated using a sliding-window approach (window length = 20, 40, 60, 80, 100 TRs). Additionally, amplitude of low-frequency fluctuation (ALFF) maps were generated using DPARSF~\cite{yan2010dparsf} to capture regional activity patterns.

\subsubsection{Diffusion Tensor Imaging Processing}

DTI data were processed using FSL v6.0 (\url{https://fsl.fmrib.ox.ac.uk/fsl}) and PANDA v1.3.1~\cite{cui2013panda}. The preprocessing steps included: 
(i) skull stripping using BET; 
(ii) eddy current and head motion correction via \texttt{eddy\_correct}; 
(iii) fitting of the diffusion tensor model using \texttt{dtifit} to derive fractional anisotropy (FA) and mean diffusivity (MD) maps; 
(iv) probabilistic tractography with \texttt{probtrackx2} using seed masks defined by the AAL template.

% Structural connectivity matrices were constructed by counting the number of streamlines between each pair of ROIs. To account for differences in cortical surface area, fiber counts were normalized by the geometric mean of the source and target region volumes, resulting in a corrected structural connectivity strength metric:
% \begin{equation}
%     S_{ij}^{\text{norm}} = \frac{N_{ij}}{\sqrt{A_i \cdot A_j}},
% \end{equation}
% where $N_{ij}$ is the raw streamline count between regions $i$ and $j$, and $A_i$, $A_j$ denote their respective cortical areas.

\subsection{Feature Construction and Input Representation}
\label{sec:features}

% The final input to \textit{NeuroAlign} consists of four complementary modalities:
% \begin{itemize}[leftmargin=*,noitemsep]
%     \item \textbf{DFC}: dynamic functional connectivity sequence ($2 \times 164 \times 164$,  $4 \times 164 \times 164$, $8 \times 164 \times 164$, $16 \times 164 \times 164$)
%     \item \textbf{SFC}: static functional connectivity matrix ($164 \times 164$)
%     \item \textbf{ALFF}: regional activity map ($64 \times 64 \times 36$ volume)
%     \item \textbf{FA}: fractional anisotropy ($64 \times 64 \times 40$ volume)
% \end{itemize}

\begin{itemize}[leftmargin=*,noitemsep]
    \item \textbf{DFC}: Multi-scale dynamic functional connectivity sequences, derived using sliding windows of increasing duration (10, 20, 30, 40, and 50 TRs, corresponding to 20, 40, 60, 80, and 100 s when TR = 2 s), each producing a time-varying $N_s \times 164 \times 164$ tensor of Pearson correlations over the Dosenbach-164 atlas.
    
    \item \textbf{SFC}: Static functional connectivity matrix ($164 \times 164$) computed as full-session average correlation between ROIs.
    
    \item \textbf{ALFF}: Amplitude of low-frequency fluctuations, mapped as a 3D volume in standardized brain space ( $64 \times 64 \times 34$, isotropic 3mm voxels).
    
    \item \textbf{FA}: Fractional anisotropy map from DTI, coregistered to fMRI space and resampled to match spatial resolution ($64 \times 64 \times 34$).
\end{itemize}
Each modality is independently projected into a latent embedding space through dedicated encoding pathways before being fed into the DMHA and DDHI modules. All features are z-score normalized per subject to mitigate site-related intensity variations.

\subsection{Experimental Setup}
\label{sec:experiment details}

\subsubsection{Implementation Details}
All implementation settings were kept identical across the five-fold cross-validation experiments. Within the \textit{DMHA} component of \textbf{\textit{NeuroAlign}}, convolution kernels were set to a size of 3 and stride of 1, except for the first layer in each modality branch and the pyramid adaptive convolution pipeline, where the stride was set to 2 for dimensionality reduction. The length of the dual-modal feature vector before being fed into DDHI was set to 512. The dimension of the feed-forward layer in the residual mixer was set to 16. The number of self-attention heads in FI and GI was set to 16.

The number of filters in this pipeline increased progressively with the temporal scale, starting from 8 and doubling to 16, 32, and 64. The \textit{DDHI} component employed a 16-head single-layer attention mechanism. We trained the model with batches of 4 and a learning rate of $10^{-4}$, using the Adam optimization strategy. Experiments were carried out on a server with an NVIDIA 2 $\times$ 4090Ti GPU, utilizing PyTorch for 2 hours. 

\subsubsection{Cross-validation and Statistical Evaluation}
\label{sec:statistical_evaluation}

All experiments were evaluated using subject-level stratified five-fold cross-validation rather than a single 8:1:1 split. 
In each fold, one subset was used as the independent test set, and the remaining subjects were used for training and validation. 
The class distribution was kept as balanced as possible across folds, and no subject appeared in both the training and testing sets. 
Accuracy, recall, precision, and F1-score were calculated for each fold. 
To provide a more transparent assessment of performance stability, we further visualized the fold-wise accuracy distributions for the compared methods. 
These fold-wise distributions were used as descriptive statistical evidence to show the variability of model performance across different data partitions.
For cross-dataset validation, the model trained in each fold on the source dataset was directly evaluated on the target dataset without using target-domain labels during training, and the reported transfer results were averaged across the five fold-specific models.

\subsubsection{Modality-Controlled Baseline Implementation}
\label{sec:modality_controlled_baselines}

\begin{table*}[htbp]
\centering
\scriptsize
\caption{Representative adaptation strategies of SOTA methods under the unified DFC+SFC+ALFF+FA input setting.}
\label{tab:baseline_adaptation}
\setlength{\tabcolsep}{4pt}
\renewcommand{\arraystretch}{1.15}

\begin{tabularx}{\textwidth}{
>{\raggedright\arraybackslash}p{2.3cm}
>{\raggedright\arraybackslash}p{3.0cm}
>{\raggedright\arraybackslash}p{3.2cm}
>{\raggedright\arraybackslash}X}
\toprule
\textbf{Representative Method} 
& \textbf{Original Focus} 
& \textbf{Preserved Core Design} 
& \textbf{Adaptation to DFC+SFC+ALFF+FA} \\
\midrule

\multicolumn{4}{l}{\textbf{Representative adapted methods in the GUTCM comparison}} \\
\midrule

Fang et al.~\cite{fang2022re} 
& Multimodal brain-image-based early AD diagnosis 
& Transfer learning and multimodal representation learning strategy 
& Four modality-specific input branches were introduced for DFC, SFC, ALFF, and FA. The original multimodal learning strategy and classifier were preserved as much as possible. \\

Gao et al.~\cite{gao2023hybrid} 
& MRI-based AD diagnosis with multi-scale feature modeling 
& Hybrid multi-scale attention convolution and transformer-based fusion 
& DFC/SFC connectivity matrices and ALFF/FA regional maps were projected into feature embeddings and then passed to the preserved multi-scale attention and transformer modules. \\

He et al.~\cite{he2024spatiotemporal} 
& rs-fMRI-based spatiotemporal graph modeling 
& Spatiotemporal graph transformer and classification head 
& DFC was used as the main dynamic graph input, while SFC, ALFF, and FA were encoded as auxiliary connectivity-level or region-level attributes. \\

Li et al.~\cite{li2025transformer} 
& Transformer-based neuroimaging representation learning 
& Transformer attention module and high-level classification framework 
& DFC, SFC, ALFF, and FA were encoded as four modality-specific tokens. The transformer attention module was retained for multimodal feature interaction. \\

\midrule
\multicolumn{4}{l}{\textbf{Representative adapted methods in the ADNI/OASIS comparison}} \\
\midrule

Jeon et al.~\cite{jeon2020enriched} 
& rs-fMRI-based representation learning for early MCI diagnosis 
& Spatiotemporal feature learning and classification framework 
& The original rs-fMRI input branch was extended to four feature-specific encoders for DFC, SFC, ALFF, and FA, followed by the preserved representation learning and classification modules. \\

Dong et al.~\cite{dong2021integration} 
& Functional connectivity-based MCI diagnosis 
& Integration of handcrafted FC features and embedded deep features 
& DFC and SFC were used to construct connectivity-level features, while ALFF and FA were encoded as regional embeddings and integrated before classification. \\

Cao et al.~\cite{cao2023novel} 
& Dynamic functional connectivity feature analysis 
& DFC-derived feature extraction and classification strategy 
& DFC remained the main input branch, while SFC, ALFF, and FA were introduced through modality-specific adapters and fused with the DFC representation. \\

Chen et al.~\cite{chen2023orthogonal} 
& Multimodal AD diagnosis 
& Orthogonal latent-space learning, feature weighting, and graph learning 
& DFC, SFC, ALFF, and FA were treated as four modality views and projected into the shared latent space while preserving the original weighting and graph-learning modules. \\

\bottomrule
\end{tabularx}

\vspace{1mm}
\parbox{\linewidth}{
\scriptsize
\textit{Note:}
This table provides representative examples of how SOTA methods were adapted under the unified DFC+SFC+ALFF+FA input setting. 
For the remaining compared methods in both the GUTCM and ADNI/OASIS comparisons, we followed the same adaptation principle: modality-specific input adapters were introduced when necessary, while the original backbone, attention mechanism, graph modeling strategy, fusion module, and classifier were preserved as much as possible. 
For methods originally designed for single-modality MRI, rs-fMRI, EEG, or non-AD tasks, the adapted version should be interpreted as an architecture-level comparison under matched multimodal inputs rather than a direct reproduction of the original experimental setting.
}
\end{table*}

For fair evaluation, we adopted a modality-controlled experimental protocol for all re-implemented baseline models. Both \textbf{\textit{NeuroAlign}} and the adapted baselines used the same set of multimodal inputs, namely DFC, SFC, ALFF, and FA. This setting was necessary because existing approaches that natively support this exact combination of dynamic functional connectivity, static functional connectivity, regional functional activity, and structural diffusion information are very limited. For example, Jeon et al.~\cite{jeon2020enriched} mainly focuses on spatiotemporal functional representations, Dong et al.~\cite{dong2021integration} primarily operates on functional connectivity features, and Cao et al.~\cite{cao2023novel} emphasizes DFC-derived feature analysis. Directly comparing these original input settings with \textbf{\textit{NeuroAlign}} would introduce modality mismatch and could confound the interpretation of performance gains. 

Therefore, to ensure that the comparison reflects differences in fusion and representation learning strategies rather than differences in available input information, we adapted the input interfaces of representative baseline models to receive the unified four-feature inputs. During this adaptation, modality-specific input adapters were introduced when necessary, while the original core network structures, fusion strategies, and classification modules of the baseline models were preserved as much as possible. All methods were trained and evaluated using the same preprocessing pipeline, training/validation/testing split, feature normalization strategy, and evaluation metrics. Accordingly, the controlled comparisons reported in this study are conducted under matched input-modality conditions.

To clarify the modality-controlled comparison protocol, Table~\ref{tab:baseline_adaptation} summarizes representative examples of how the compared SOTA methods were adapted to the unified DFC+SFC+ALFF+FA input setting. 
The examples are grouped according to their use in the GUTCM comparison and the ADNI/OASIS comparison. 
For the remaining methods, the same adaptation principle was applied: the input interfaces were modified to receive four multimodal features, while the original core model structures were preserved as much as possible.

\subsection{Total Loss}
\label{sec: loss}
To effectively map cognitive conditions (e.g., MCI, SCD, etc.), the residual projector combines the embeddings \( \text{G}_{C} \) and \( \text{G}_{R} \) (C: Connectivity; R: Regional), reducing the dimension to match the number of classes. The classification loss \( \mathcal{L}_{CLS} \) is estimated using cross-entropy, {ensuring the model learns representations related to cognitive decline for classification from DDHI and DMHA. During the back-propagation process, we utilized the summarized loss as defined in Eq.
(\ref{eq: Loss}).
    \begin{equation}
    \label{eq: Loss}
        \mathcal{L} = \mathcal{L}_{CLS} + \mathcal{L}_{DSA} +  \mathcal{L}_{FSA}
    \end{equation}

%We validated our proposed \textbf{\textit{NeuroAlign}} framework at four distinct levels: 
%(1) Outperformance validation, comparing it against other existing methods; 
%(2) Generalizability validation, using both public ADNI and in-house GUTCM datasets;
%(3) Ablation study from the perspective of modalities; 
%(4) Ablation study focusing on the architectural aspects.

%To quantify the framework's performance, metrics including accuracy, recall, precision, and F1-score were employed.
%For details on the first two levels of validation, refer to Section \ref{sec:5.1}. Information regarding the ablation studies can be found in Section \ref{sec:5.2}. Additionally, the explanation results derived from our introduced \textbf{\textit{SAM}} technique are presented in Section \ref{sec:5.3}.

% \begin{enumerate}[label=$\bullet$]
%     \setlength\itemsep{0em}
%     \setlength\parsep{0em}
%     \item Directly connecting a ResNet model to a Fully Connected Layer (FL) ($R_{*}$FL).
%     \item Applying an attention mechanism to channels before forwarding to the FL for classification ($R_{*}$AFL).
%     \item Directly connecting a ResNet model to the transformer blocks ($R_{*}$T).
%     \item Using features with applied channel attention as input for the transformer blocks ($R_{*}$AT).
% \end{enumerate}

    % \begin{tabularx}{\textwidth}{>{\hsize=1.5\hsize\raggedright\arraybackslash}X*{9}{>{\hsize=.5\hsize\centering\arraybackslash}X}}

% \vspace{-5mm}

\begin{table*}[htbp]
    \centering
    \scriptsize
    \caption{Controlled comparison between the proposed \textbf{\textit{NeuroAlign}} and adapted SOTA models on the GUTCM dataset}
    \label{tab:gutcm_comparison}
    \resizebox{\linewidth}{!}{
    \begin{tabularx}
    {\textwidth}{c*{9}{>{\centering\arraybackslash}X}}
        \toprule
         & Fang et al., \cite{fang2022re} & Gao et al., \cite{gao2023hybrid} & Zhao et al., \cite{zhao2023depression} & Pradhan et al., \cite{pradhan2024advance} & He et al., \cite{he2024spatiotemporal} & Cseker et al., \cite{cseker2025investigating} & Francis et al., \cite{francis2025deep} & Li et al., \cite{li2025transformer} & \textbf{\textit{NA}} \\
        \midrule
        \textbf{SCD/HC} \\
        Accuracy & 0.791 & 0.895 & 0.766 & 0.895 & 0.896 & 0.851 & 0.912 & 0.779 & \textbf{0.950} \\
        Recall & \textbf{1.000} & 1.000 & 1.000 & 0.862 & 0.946 & 1.000 & 0.887 & 1.000 & 0.950 \\
        Precision & 0.791 & 0.895 & 0.766 & 1.000 & 0.936 & 0.841 & 1.000 & 0.779 & \textbf{0.950} \\
        F1-score & 0.876 & 0.924 & 0.865 & 0.924 & 0.934 & 0.909 & 0.937 & 0.873 & \textbf{0.950} \\
        \midrule
        \textbf{MCI/SCD} \\
        Accuracy & 0.750 & 0.562 & 0.625 & 0.562 & 0.625 & 0.688 & 0.688 & 0.656 & \textbf{0.889} \\
        Recall & 0.694 & 1.000 & 0.338 & 1.000 & 0.613 & 0.562 & 0.450 & 0.650 & 0.800 \\
        Precision & 0.791 & 0.562 & 0.750 & 0.562 & 0.750 & 0.852 & 1.000 & 0.667 & \textbf{1.000} \\
        F1-score & 0.730 & 0.718 & 0.440 & 0.718 & 0.650 & 0.667 & 0.604 & 0.653 & \textbf{0.889} \\
        \midrule
        \textbf{MCI/HC} \\
        Accuracy & 0.562 & 0.656 & 0.562 & 0.656 & 0.545 & 0.545 & 0.697 & 0.562 & \textbf{0.812} \\
        Recall & 1.000 & 0.562 & 1.000 & 0.562 & 1.000 & 1.000 & 1.000 & \textbf{1.000} & 0.900 \\
        Precision & 0.562 & 0.729 & 0.562 & 0.729 & 0.562 & 0.545 & 0.664 & 0.606 & \textbf{0.818} \\
        F1-score & 0.718 & 0.624 & 0.718 & 0.624 & 0.718 & 0.694 & 0.732 & 0.718 & \textbf{0.857} \\
        \midrule
        \textbf{MCI/SCD/HC} \\
        Accuracy & 0.574 & 0.516 & 0.532 & 0.516 & 0.625 & 0.652 & 0.489 & 0.457 & \textbf{0.700} \\
        Recall & 1.000 & 1.000 & \textbf{1.000} & 1.000 & 0.710 & 0.707 & 1.000 & 0.800 & 0.950 \\
        Precision & 0.656 & 0.617 & 0.632 & 0.617 & 0.900 & \textbf{0.942} & 0.612 & 0.606 & 0.850 \\
        F1-score & 0.779 & 0.741 & 0.754 & 0.741 & 0.794 & 0.791 & 0.745 & 0.663 & \textbf{0.880} \\
        \bottomrule
    \end{tabularx}
    }

    \vspace{1mm}
    \parbox{\linewidth}{
    \scriptsize
    \textit{Note:} 
 All compared SOTA models were re-implemented using the same multimodal inputs as \textbf{\textit{NeuroAlign}}: DFC, SFC, ALFF, and FA. 
Because few existing methods natively support this exact four-feature setting, their input stages were adapted while preserving the original core architectures and classifiers as much as possible. 
Thus, this comparison is modality-controlled and mainly reflects differences in multimodal fusion strategies rather than input modalities.
    }
\end{table*}

\section{Results}

\begin{figure}
  \centering
  \includegraphics[width=0.99\linewidth]{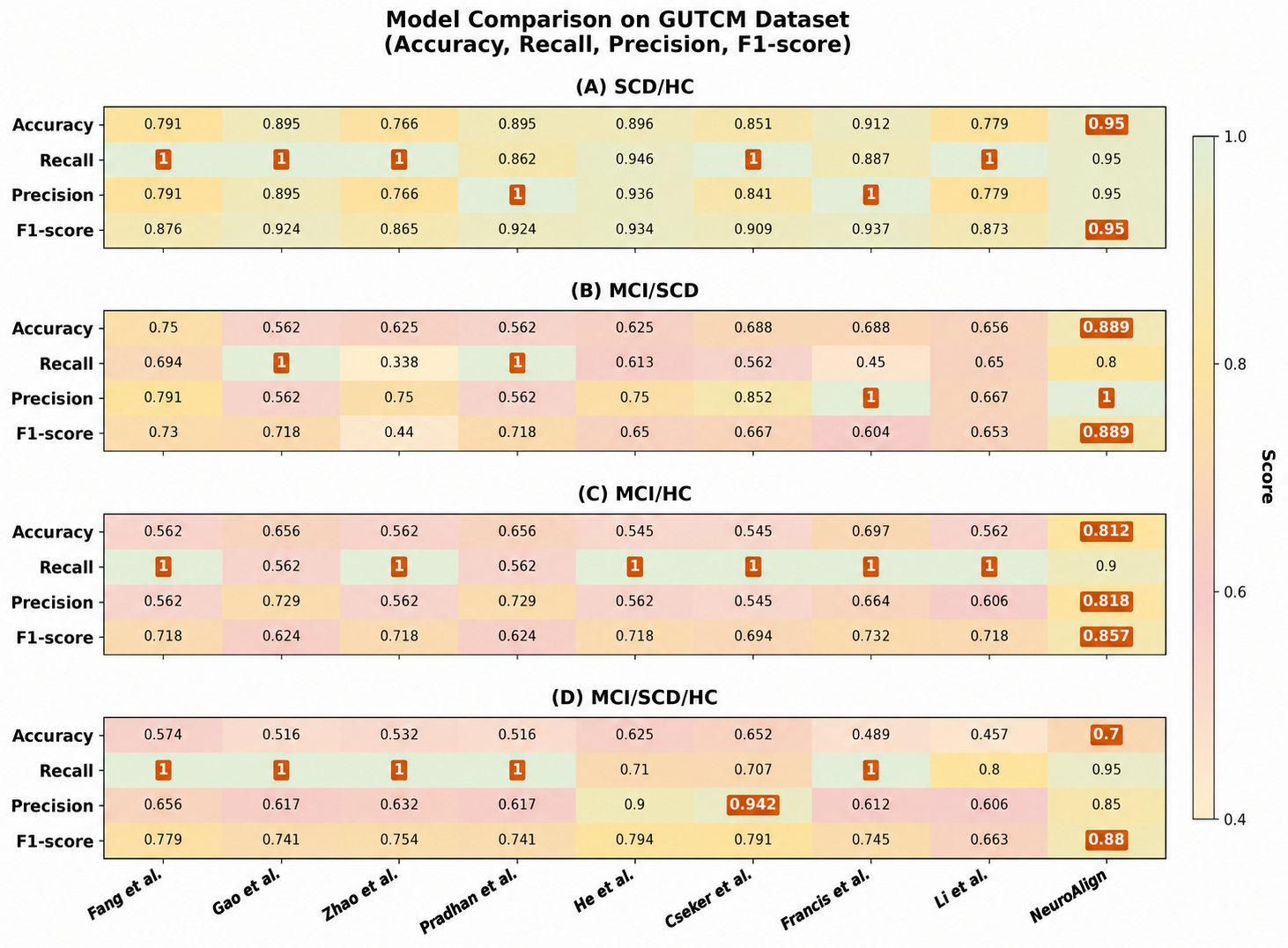}
  \caption{Comparison of NeuroAlign against SOTA methods on the GUTCM dataset. 
    Subfigures (A--D) correspond to diagnostic tasks: (A) SCD/HC, (B) MCI/SCD, (C) MCI/HC, and (D) MCI/SCD/HC. 
    Each cell shows the score (Accuracy, Recall, Precision, or F1-score) of a model on that task-metric pair. 
    Best scores per metric are highlighted in orange..}
  \label{fig:GUTCM}
\end{figure}
% In the following sections, we present a comprehensive evaluation of the proposed \textbf{\textit{NeuroAlign}} framework through comparative analysis against existing methodologies. Our experiments are conducted on both our in-house GUTCM dataset (\textit{cf. Section~\ref{sec:5.1.1}}) and the publicly available ADNI and OASIS datasets (\textit{cf. Section~\ref{sec:5.1.2}}). We further provide visual interpretations of our Synergistic Activation Mapping (SAM)}, including learned connectivity patterns and region-wise attention maps (\textit{cf. Section~\ref{sec:5.2}}). Following the presentation of quantitative comparisons and qualitative explainability results, we perform an ablation study to validate the effectiveness of our alignment strategy (\textit{cf. Section~\ref{sec:5.3}}). Finally, to assess the generalizability of \textit{NeuroAlign}, we conduct cross-dataset validation experiments, demonstrating its robustness across heterogeneous data sources (\textit{cf. Section~\ref{sec:5.4}}).

We evaluate \textbf{\textit{NeuroAlign}} on the GUTCM, ADNI, and OASIS datasets to assess its classification performance, feature-level attribution behavior, and preliminary cross-dataset transferability in cognitive impairment analysis. 
The experiments are organized from controlled within-dataset evaluation to external reference comparison and cross-dataset validation. 
First, we compare \textbf{\textit{NeuroAlign}} with adapted SOTA models on the GUTCM dataset under the unified DFC+SFC+ALFF+FA input setting, so that the comparison mainly reflects differences in multimodal fusion strategies rather than differences in available input modalities. 
Second, we report dataset-specific reference comparisons on the public ADNI and OASIS datasets to examine how the proposed framework performs relative to previously reported methods under comparable diagnostic tasks. 
Third, we conduct hierarchical ablation studies to evaluate the contributions of DMHA, DDHI, and the alignment objectives used in DSA and FSA. 
Finally, we provide SAM-based attribution analysis and cross-dataset transfer experiments to inspect modality-specific feature usage and remaining domain gaps across heterogeneous acquisition settings.

\begin{table*}[htbp]
    \centering
    \scriptsize
    \caption{Performance comparison on ADNI and OASIS datasets. SOTA methods are listed as dataset-specific reference comparisons, and `--' denotes unavailable results.}
    \label{tab:combined_comparison}
    \setlength{\tabcolsep}{4pt}
    \renewcommand{\arraystretch}{1.15}
    \begin{tabularx}{\textwidth}{l *{8}{>{\centering\arraybackslash}X}}
        \toprule
        \multirow{2}{*}{\textbf{Methods}} &
        \multicolumn{4}{c}{\textbf{ADNI}} &
        \multicolumn{4}{c}{\textbf{OASIS}} \\
        \cmidrule(lr){2-5} \cmidrule(lr){6-9}
        & \textbf{Acc} & \textbf{Rec} & \textbf{Pre} & \textbf{F1} 
        & \textbf{Acc} & \textbf{Rec} & \textbf{Pre} & \textbf{F1} \\
        \midrule
        \multicolumn{9}{l}{\textbf{SOTA Methods (Dataset-Specific)}} \\
        \midrule
        \cite{lei2021auto}              & 0.822 & 0.829 & 0.784    & 0.806 & --    & --    & --    & --    \\
        \cite{jeon2020enriched}         & 0.705 & 0.729 & 0.763   & 0.746   & --    & --    & --    & --    \\
        \cite{dong2021integration}      & 0.784 & 0.815    & 0.819    &  0.817    & --    & --    & --    & --    \\
        \cite{cao2023novel}             & 0.820 & 0.840 & 0.800 & 0.830 & --    & --    & --    & --    \\
        \cite{chen2023orthogonal}       & 0.671 & 0.697 & 0.734   &  0.715    & --    & --    & --    & --    \\
        \cite{gao2023hybrid}            & 0.730 & 0.750 & 0.780 & 0.765 & --    & --    & --    & --    \\
        \cite{fang2022re}               & 0.590 & 0.900 & 0.640 & 0.710 & 0.563 & 0.667 & 0.556 & 0.606 \\
        \cite{cseker2025investigating}  & 0.730 & 0.750 & 0.780 & 0.765 & 0.672 & 0.689 & 0.660 & 0.674 \\
        \cite{yu2023multi}              & 0.600 & 0.400 & 0.400 & 0.400 & 0.561 & 0.711 & 0.552 & 0.621 \\
        \cite{yu2024rdgt}               & 0.600 & 0.600 & 0.600 & 0.600 & 0.574 & 0.733 & 0.559 & 0.634 \\
        \cite{gao2023multimodal}        & --    & --    & --    & --    & 0.538 & 0.600 & 0.529 & 0.562 \\
        \cite{carcagni2023convolution}  & --    & --    & --    & --    & 0.548 & 0.644 & 0.537 & 0.586 \\
        \cite{huang2024ensemble}        & --    & --    & --    & --    & 0.603 & 0.711 & 0.582 & 0.640 \\
        \midrule
        \textbf{\textit{NeuroAlign}}    
        & \textbf{0.825} & \textbf{0.900} & 0.780 & \textbf{0.836} 
        &  \textbf{0.735} & \textbf{0.800} & \textbf{0.735} & \textbf{0.766}  \\
        \bottomrule
    \end{tabularx}

    \vspace{1mm}
    \parbox{\linewidth}{
    \scriptsize
    \textit{Note:}
    This table provides dataset-specific reference comparisons with previously reported SOTA results on ADNI and OASIS. 
    Since these published methods may differ in input modalities, feature representations, and experimental protocols, this table is used as an external benchmark. 
    The modality-controlled comparison with adapted baselines under the unified DFC+SFC+ALFF+FA input setting is provided in Table~\ref{tab:gutcm_comparison}.
    }
\end{table*}

\begin{figure}
  \centering
  \includegraphics[width=0.98\linewidth]{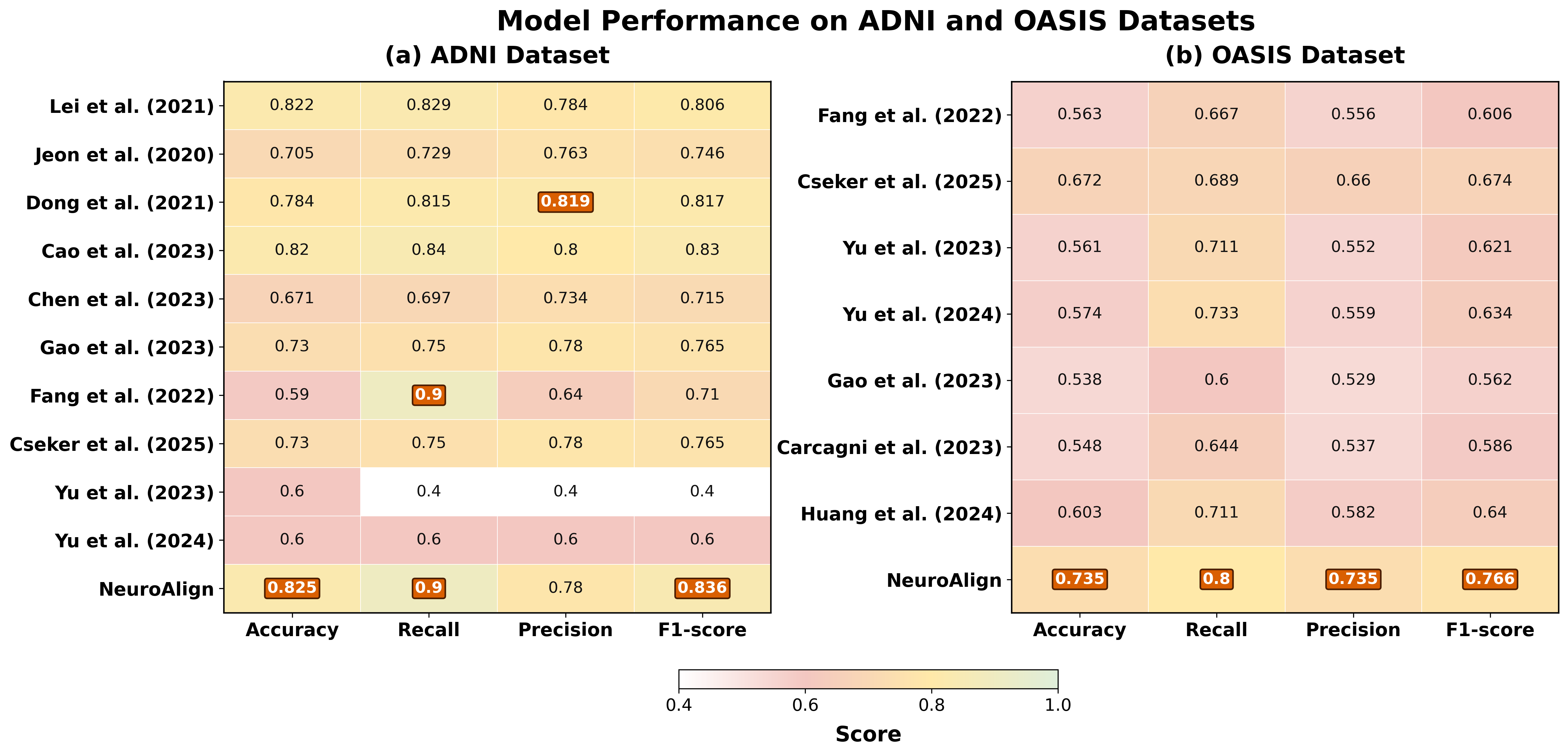}
  \caption{Performance comparison on (a) ADNI and (b) OASIS datasets. 
    Model names include publication years; \textit{NeuroAlign} is shown without year. 
    '--' denotes missing results. Best scores per metric are highlighted in orange; 
    \textit{NeuroAlign}'s results are outlined in black.}
  \label{fig:public}
\end{figure}

% For this comparison, we selected state-of-the-art (SOTA) methods [\cite{chen2023orthogonal,cseker2025investigating,huang2024ensemble,lei2021auto},e.g.,]that have been previously applied to the ADNI dataset for identifying MCI from HC patterns to benchmark against \textbf{\textit{NeuroAlign}}. 

\subsection{Comparison Study}

\subsubsection{Baseline Comparison on the GUTCM dataset}
\label{sec:5.1.1}

Additionally, to validate the effectiveness of the proposed framework under a strictly controlled multimodal setting, we constructed adapted SOTA comparison models based on existing algorithms that share architectural or methodological similarities with \textbf{\textit{NeuroAlign}}~[\cite{gao2023hybrid,he2024spatiotemporal,faghiri2020weighted,li2025transformer}]. 
For all compared methods, the input interfaces were adapted to receive the same four neuroimaging features, including DFC, SFC, ALFF, and FA, while their original backbone structures, attention mechanisms, fusion strategies, and classification modules were preserved as much as possible. 
Therefore, the comparison on the GUTCM dataset was conducted under a modality-controlled setting, where performance differences mainly reflect the effectiveness of multimodal fusion and representation learning strategies rather than differences in available input information.

As shown in Table~\ref{tab:gutcm_comparison} and Fig.~\ref{fig:GUTCM}, \textbf{\textit{NeuroAlign}} achieves higher overall performance than the adapted comparison models in several diagnostic tasks. 
In particular, the proposed method obtains consistent improvements in accuracy and F1-score for the SCD/HC, MCI/SCD, MCI/HC, and three-class MCI/SCD/HC classification tasks. 
These results suggest that the hierarchical design of \textit{DMHA} and \textit{DDHI} is useful for integrating heterogeneous connectivity-level and region-level features. 
Specifically, \textit{DMHA} helps reduce representational mismatch among multi-scale DFC, SFC, ALFF, and FA features, whereas \textit{DDHI} further promotes fine-grained and global interactions between connectivity-domain and region-domain information. 
Compared with direct feature concatenation or conventional attention-based fusion, this structured alignment-and-interaction strategy provides a more effective way to exploit complementary functional and structural information.

In contrast to the relatively marginal gain observed on ADNI, the GUTCM results show more pronounced improvements across multiple diagnostic tasks, especially for the more challenging MCI/SCD and three-class MCI/SCD/HC settings. 
These tasks involve subtler cognitive transitions and more heterogeneous disease-related patterns than the binary MCI/HC classification task, making them more sensitive to the quality of multimodal representation learning. 
The stronger performance of \textbf{\textit{NeuroAlign}} on these tasks indicates that the benefit of hierarchical alignment and interaction becomes more evident when the model is required to distinguish adjacent or intermediate cognitive states. 
This finding supports the motivation of explicitly modeling dynamic--static, functional--structural, and connectivity--regional relationships, rather than relying on unstructured multimodal fusion.

\subsubsection{Reference Comparison on the ADNI and OASIS Datasets}
\label{sec:5.1.2}

For more comparison, we selected SOTA methods [\cite{chen2023orthogonal,cseker2025investigating,huang2024ensemble,lei2021auto},e.g.,] that have been previously applied to the ADNI and OASIS dataset for identifying MCI from HC patterns to benchmark against \textbf{\textit{NeuroAlign}}. Our method have higher ACC and F1-score in the ADNI and OASIS dataset compared with the SOTA methods, demonstrating our method have enhanced roundness in diagnosis negative subjects (HCs) and positive subjects (MCIs/SCDs).

Compared with previously reported SOTA results on ADNI and OASIS, \textbf{\textit{NeuroAlign}} achieves competitive or slightly higher overall performance. However, these external comparisons may involve differences in preprocessing, input features, and evaluation protocols; therefore, they should be interpreted together with the modality-controlled comparisons and ablation studies.

Although \textbf{\textit{NeuroAlign}} achieves the highest accuracy on the ADNI dataset, we note that the improvement over the strongest reported method, Cao et al.~\cite{cao2023novel}, is numerically small in terms of accuracy. Therefore, we do not interpret this marginal accuracy gain alone as sufficient evidence for the superiority of the proposed framework. Instead, the value of \textbf{\textit{NeuroAlign}} should be assessed from multiple perspectives, including balanced classification performance, consistency across datasets, modality-controlled comparisons, and component-level ablation analysis.

Specifically, compared with Cao et al.~\cite{cao2023novel}, \textbf{\textit{NeuroAlign}} shows a larger improvement in recall on ADNI, indicating improved sensitivity to MCI subjects, while maintaining comparable F1-score. On OASIS, where available SOTA results are generally lower, \textbf{\textit{NeuroAlign}} achieves more evident improvements in both accuracy and F1-score. These results suggest that the proposed method does not merely optimize a single metric on one dataset, but tends to provide more balanced performance across heterogeneous cohorts. Nevertheless, we acknowledge that the absolute accuracy gain on ADNI is limited, and thus the comparison with previously reported SOTA methods should be regarded as reference evidence rather than the sole validation of the proposed architecture.

After expanding the OASIS subset under the paired multimodal inclusion criteria, \textbf{\textit{NeuroAlign}} achieves 0.735 accuracy, 0.800 recall, 0.735 precision, and 0.766 F1-score on OASIS. The enlarged OASIS cohort provides a more stable external evaluation setting. Nevertheless, because the OASIS subset remains smaller than GUTCM and ADNI, the OASIS results are still interpreted as preliminary external validation rather than conclusive evidence of large-scale generalizability. Larger independent cohorts are still required to further verify the stability of the proposed framework.

\begin{figure}
  \centering
  \includegraphics[width=0.95\linewidth]{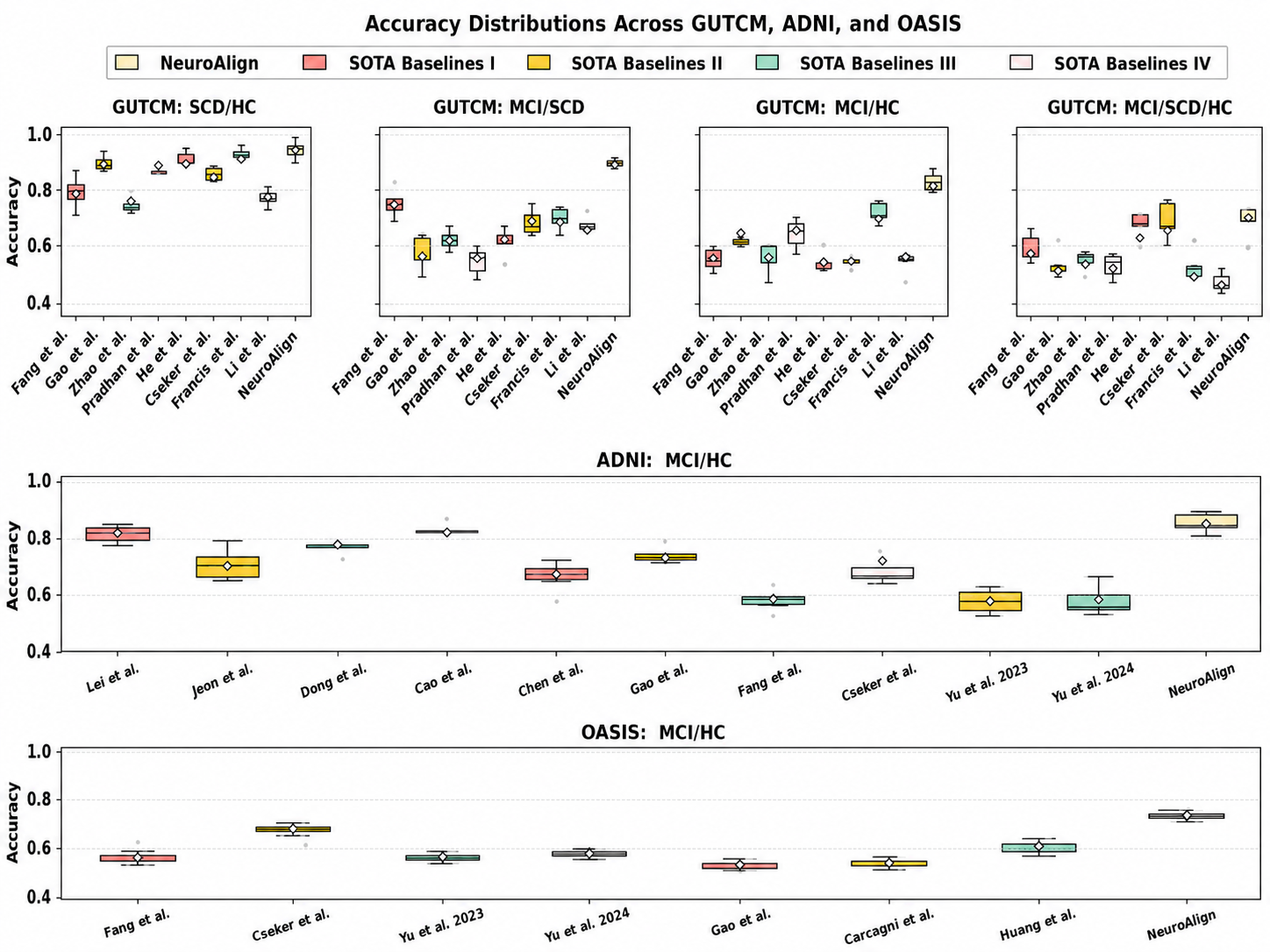}
  \caption{Statistical visualization of accuracy distributions across five-fold cross-validation. 
  The first row presents the modality-controlled comparison on the GUTCM dataset for four diagnostic tasks: SCD/HC, MCI/SCD, MCI/HC, and MCI/SCD/HC. 
  The second and third rows show the dataset-specific reference comparisons on ADNI and OASIS, respectively, for the MCI/HC task. 
  In each panel, boxplots summarize the fold-wise accuracy distributions of different methods, and diamond markers indicate the accuracy  reported in the corresponding tables. 
  \textit{NeuroAlign} is highlighted in pale yellow, while the other methods are grouped using consistent pastel color families for visual comparison.}
  \label{fig:statics}
\end{figure}

\subsubsection{Statistical Visualization of Performance}

To provide a more intuitive view of the performance stability across different methods and datasets, we further visualized the accuracy distributions under the five-fold cross-validation protocol in Fig.~\ref{fig:statics}. 
For the GUTCM dataset, the first row shows the fold-wise accuracy distributions for the four classification tasks, including SCD/HC, MCI/SCD, MCI/HC, and MCI/SCD/HC, under the modality-controlled setting. 
For the public datasets, the second and third rows present the corresponding accuracy distributions on ADNI and OASIS for the MCI/HC task, respectively. 
In each panel, the boxplots summarize the distribution of fold-wise results, while the diamond markers denote the accuracy values reported in Tables~\ref{tab:gutcm_comparison} and \ref{tab:combined_comparison}. 
Overall, \textbf{\textit{NeuroAlign}} shows consistently competitive or superior median accuracy across the evaluated tasks, with relatively stable fold-wise variations.

These visualizations complement the numerical comparisons by showing the variability of fold-wise performance rather than only reporting average values. The ADNI and OASIS panels should be interpreted as dataset-specific reference comparisons, since the included SOTA methods may differ in modalities and protocols. In contrast, the GUTCM panels provide a modality-controlled comparison under the unified DFC+SFC+ALFF+FA input setting. The results suggest that \textbf{\textit{NeuroAlign}} maintains relatively stable performance across folds, although its advantage on public datasets should still be interpreted cautiously due to differences in experimental settings.

\begin{table}[htbp]
\caption{SAM-based DFC attribution maps in the connectivity domain. The maps summarize model-relevant dynamic connectivity patterns under single-feature and multi-feature settings and should be interpreted as model-derived attribution rather than validated clinical biomarkers.}
\label{tab:Table11}
\centering
\scriptsize
\renewcommand{\arraystretch}{1.05}
\setlength{\tabcolsep}{4pt}
\begin{tabularx}{\textwidth}{
>{\centering\arraybackslash}p{1.4cm}
>{\raggedright\arraybackslash}X
>{\raggedright\arraybackslash}p{1.8cm}
>{\raggedright\arraybackslash}X
>{\raggedright\arraybackslash}p{1.5cm}}
\toprule
\multirow{2}{*}{\textbf{Inputs}} & 
\multicolumn{2}{c}{\textbf{Conn (ROI Pair / Net)}} & 
\multicolumn{2}{c}{\textbf{ROIs (Region / Net)}} \\
\cmidrule(lr){2-3} \cmidrule(l){4-5}
 & \textbf{Pair} & \textbf{Net} & \textbf{ROI} & \textbf{Net} \\
\midrule
\multirow{5}{*}{Single} 
& Inf Cer--Inf Cer & CER--CER & Medial Cerebellum & CER \\
& Parietal--Parietal & SEN--SEN & Inferior Cerebellum & CER \\
& Temporal--Temporal & SEN--SEN & Anterior Insula & CON \\
& Parietal--Parietal & SEN--SEN & Anterior Insula & CON \\
& Parietal--Parietal & SEN--SEN & Posterior Insula & SEN \\
\midrule

\multirow{5}{*}{Multi} 
& PCC--PCC & DMN--DMN & Angular Gyrus & DMN \\
& PCC--Precuneus & DMN--DMN & Angular Gyrus & DMN \\
& Ang Gyrus--IPS & DMN--DMN & Dorsolateral Prefrontal Cortex & FPN \\
& vlPFC--dlPFC & FPN--FPN & Dorsal Frontal Cortex & FPN \\
& PCC--PCC & EMO--EMO & Inferior Parietal Lobule & FPN \\
\midrule

\multirow{5}{*}{$DFC(s_0)$} 
& PCC--Precuneus & DMN--DMN & Precuneus & DMN \\
& PCC--Precuneus & DMN--DMN & Precuneus & DMN \\
& PCC--Precuneus & DMN--DMN & Angular Gyrus & DMN \\
& Precuneus--Precuneus & DMN--DMN & Angular Gyrus & DMN \\
& PCC--Precuneus & DMN--DMN & Inferior Parietal Lobule & FPN \\
\midrule

\multirow{5}{*}{$DFC(s_1)$} 
& PCC--Precuneus & DMN--DMN & Precuneus & DMN \\
& PCC--Precuneus & DMN--DMN & Precuneus & DMN \\
& PCC--Precuneus & DMN--DMN & Angular Gyrus & DMN \\
& Precuneus--Precuneus & DMN--DMN & Angular Gyrus & DMN \\
& PCC--Precuneus & DMN--DMN & Inferior Parietal Lobule & FPN \\
\midrule

\multirow{5}{*}{$DFC(s_2)$} 
& vlPFC--dlPFC & FPN--FPN & Parietal Cortex & CON \\
& IPL--IPL & FPN--FPN & Parietal Cortex & CON \\
& IPL--IPL & FPN--FPN & Dorsolateral Prefrontal Cortex & FPN \\
& Occipital--Occipital & OCC--OCC & Dorsal Frontal Cortex & FPN \\
& Occipital--Post Occ & OCC--OCC & Inferior Parietal Lobule & FPN \\
\midrule

\multirow{5}{*}{$DFC(s_3)$} 
& PCC--Precuneus & DMN--DMN & Temporoparietal Junction & DMN \\
& Precuneus--Precuneus & DMN--DMN & Angular Gyrus & DMN \\
& PCC--Precuneus & DMN--DMN & Dorsolateral Prefrontal Cortex & FPN \\
& vlPFC--dlPFC & FPN--FPN & Angular Gyrus & DMN \\
& PCC--PCC & DMN--DMN & Dorsal Frontal Cortex & FPN \\
\bottomrule
\end{tabularx}
\end{table}

\subsection{Single- and Multi-modal Attribution Analysis}
\label{sec:sam_results}

To inspect the behavior of the trained model, we applied SAM to the four input representations, including DFC, SFC, ALFF, and FA. The purpose of this analysis is not to claim clinically validated biomarkers or causal pathological mechanisms. Instead, SAM is used to summarize model-derived attribution patterns in connectivity-level and region-level feature spaces.

\subsubsection{Connectivity-domain attribution}
\label{sec:connectivity_attribution}

For connectivity-domain interpretation, DFC and SFC attribution maps were visualized as ROI-to-ROI connectivity patterns and summarized by the most emphasized regions or networks. As shown in Fig.~\ref{fig: Synergy Effects-DFC}, DFC-related attribution mainly highlights dynamic connectivity patterns involving parietal, temporal, inferior cerebellar, dorsolateral prefrontal, angular gyrus, precuneus, and posterior cingulate regions. These patterns indicate which dynamic connectivity features were emphasized by the trained model under single-feature and multi-feature settings.

\begin{figure}
\centering
\includegraphics[width=0.99\textwidth]{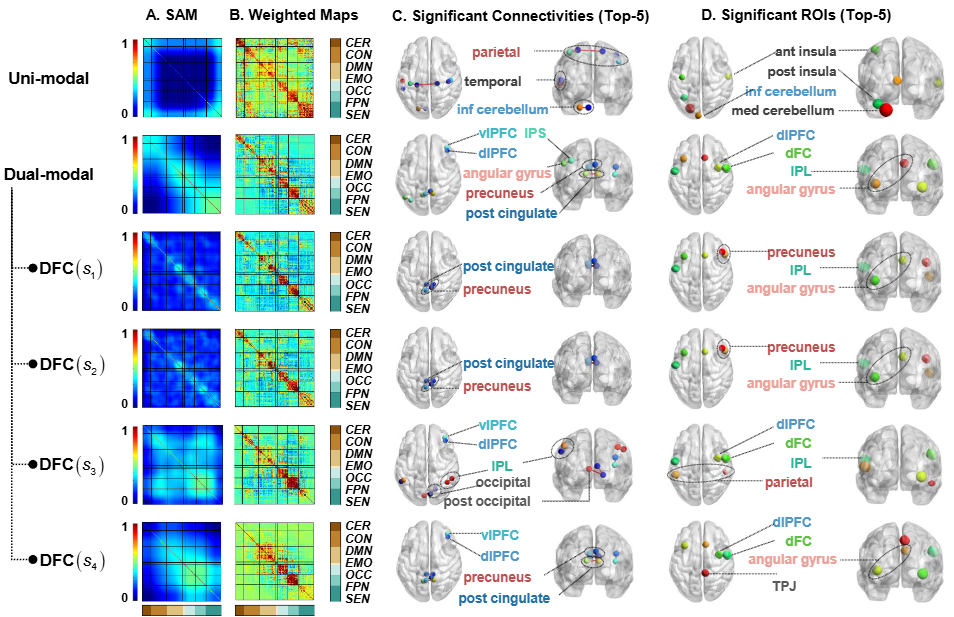}
\caption{SAM-based DFC attribution maps in the connectivity domain. The maps summarize model-relevant dynamic connectivity patterns under single-feature and multi-feature settings. Highlighted connections and regions are interpreted as model-derived attribution patterns rather than validated clinical biomarkers.}
\label{fig: Synergy Effects-DFC}
\end{figure}

\begin{table*}[htbp]
\caption{SAM-based SFC attribution maps in the connectivity domain. The maps summarize model-relevant static connectivity patterns under single-feature and multi-feature settings and should be interpreted as model-derived attribution rather than validated clinical biomarkers.}
\label{tab:Table10_small}
\centering
\scriptsize
\renewcommand{\arraystretch}{1.05}
\setlength{\tabcolsep}{4pt}
\begin{tabularx}{\textwidth}{
>{\centering\arraybackslash}p{1.4cm}
>{\raggedright\arraybackslash}X
>{\raggedright\arraybackslash}p{1.8cm}
>{\raggedright\arraybackslash}X
>{\raggedright\arraybackslash}p{1.5cm}}
\toprule
\multirow{2}{*}{\textbf{Inputs}} & 
\multicolumn{2}{c}{\textbf{Conn (ROI Pair / Net)}} & 
\multicolumn{2}{c}{\textbf{ROIs (Region / Net)}} \\
\cmidrule(lr){2-3} \cmidrule(l){4-5}
 & \textbf{Pair} & \textbf{Net} & \textbf{ROI} & \textbf{Net} \\
\midrule
\multirow{3}{*}{Single} 
& Lat Cer--Lat Cer & CER--CER & Lateral Cerebellum & CER \\
& Inf Cer--Inf Cer & CER--CER & Inferior Cerebellum & CER \\
& Med Cer--Lat Cer & CER--CER & Medial Cerebellum & CER \\
\midrule
\multirow{3}{*}{Multi} 
& Thalamus--Thalamus & CON--CON & Thalamus & CON \\
& Basal G.--Mid Insula & CON--CON & Middle Insula & CON \\
& SMA--vFC & SMA--CON & Supplementary Motor Area & SMA \\
\bottomrule
\end{tabularx}
\end{table*}

SFC attribution maps show relatively stable connectivity-level patterns involving cerebellar, parietal, angular gyrus, inferior parietal lobule, and temporo-parietal junction regions, as shown in Fig.~\ref{fig: Synergy Effects-SFC}. Compared with DFC, SFC emphasizes more stable ROI-to-ROI associations. The comparison between DFC and SFC suggests that dynamic and static functional connectivity provide complementary evidence to the classifier, but these visual patterns should be interpreted as qualitative model behavior rather than direct evidence of disease mechanisms.

\begin{figure}
\centering
\includegraphics[width=0.7\textwidth]{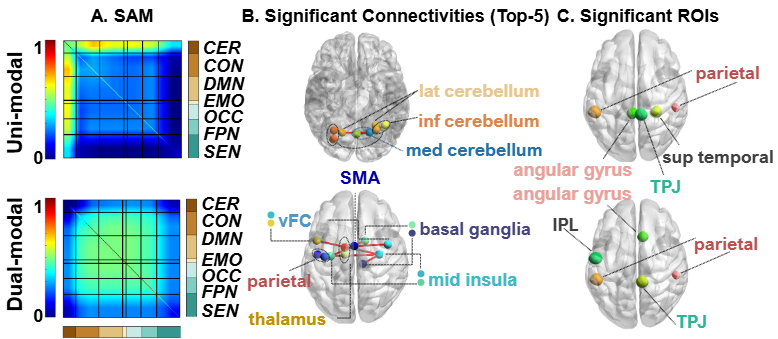}
\caption{SAM-based SFC attribution maps in the connectivity domain. The visualized connectivities indicate model-relevant static connectivity patterns under single-feature and multi-feature settings.}
\label{fig: Synergy Effects-SFC}
\end{figure}

\subsubsection{Region-domain attribution}
\label{sec:regional_attribution}

For region-domain interpretation, ALFF and FA attribution maps were visualized in the brain volume or atlas-defined regional space. As shown in Fig.~\ref{fig: Synergy Effects-Region alff}, ALFF-related attribution emphasizes model-relevant functional activity patterns around medial superior frontal, supplementary motor, putamen, and precuneus-related regions. FA-related attribution highlights structural diffusion-related patterns involving superior frontal, precentral, supplementary motor, putamen, and frontal regions. These maps indicate that the model uses complementary regional functional and structural information in addition to connectivity-level features.

\begin{figure}
\centering
\includegraphics[width=0.75\textwidth]{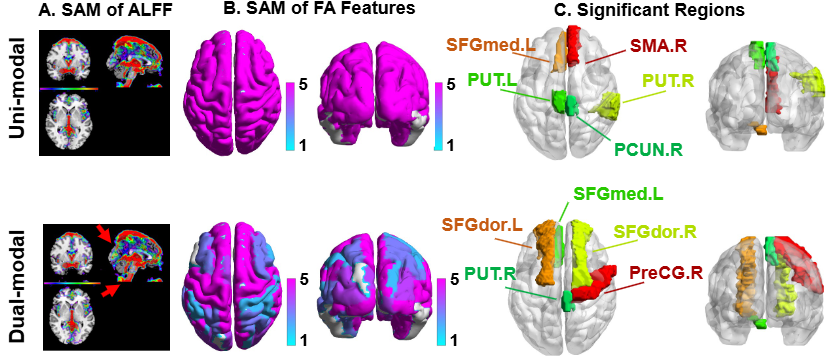} 
\caption{SAM-based regional attribution maps for ALFF and FA features. ALFF maps indicate model-relevant regional functional activity patterns, whereas FA maps indicate structural diffusion-related attribution patterns. The highlighted regions provide feature-level evidence for model inspection and should not be interpreted as causal clinical biomarkers.}
\label{fig: Synergy Effects-Region alff}
\end{figure}

% \begin{table}[htbp]%
% \caption{Representative FA attribution patterns summarized from SAM visualizations.}
% \centering
% \scriptsize
% \label{tab: Table9}
% \begin{tabularx}{\columnwidth}{XXX}
% \toprule
% Inputs & Regions & Networks \\
% \midrule
% \multirow{5}{*}{Single-feature} & SMA.R & Frontal \\
% & SFGmed.L & Prefrontal \\
% & PCUN.R & Parietal \\
% & PUT.L & Subcortical \\
% & PUT.R & Subcortical \\
% \midrule
% \multirow{5}{*}{Multi-feature} & PreCG.R & Frontal \\
% & SFGdor.R & Prefrontal \\
% & SFGdor.L & Prefrontal \\
% & SFGmed.L & Prefrontal \\
% & PUT.R & Subcortical \\
% \bottomrule
% \end{tabularx}
% \end{table}

\begin{figure}
\centering
\includegraphics[width=0.7\textwidth]{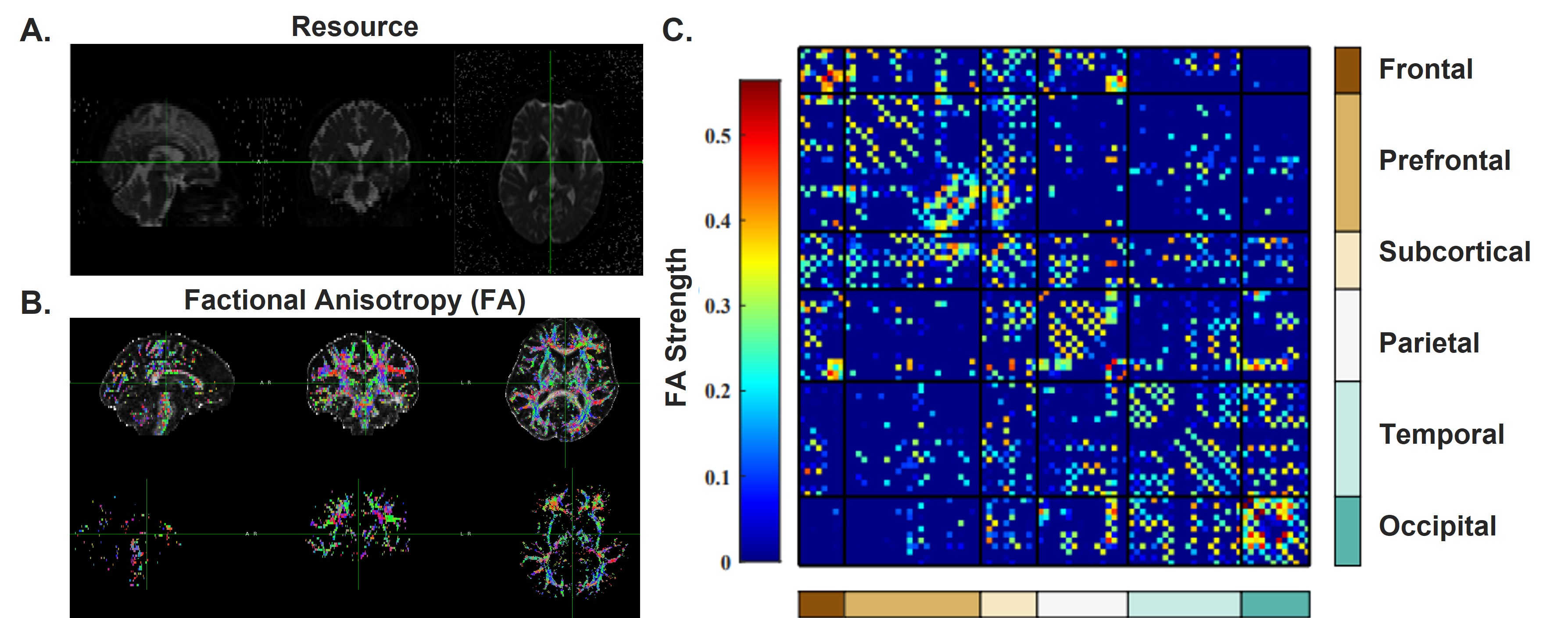}
\caption{SAM-based attribution patterns in FA-derived connectivity. A--B show the DTI-derived FA maps, and C shows the connectivity-level attribution generated by SAM.}
\label{fig: Synergy Effects-Region connect}
\end{figure}

\begin{table*}[htbp]
\centering
\caption{FA attribution patterns and modality ablation results. 
Left: representative FA attribution patterns summarized from SAM visualizations. 
Right: performance comparison of individual MRI modalities and fused multimodal inputs with our NeuroAlign.}
\label{tab:fa_attribution_modality_ablation}
\scriptsize
\renewcommand{\arraystretch}{1.15}
\setlength{\tabcolsep}{5pt}

\begin{minipage}[t]{0.48\textwidth}
\centering
\textbf{(A) Representative FA attribution patterns}
\vspace{1mm}

\begin{tabularx}{\linewidth}{
>{\centering\arraybackslash}X
>{\centering\arraybackslash}X
>{\centering\arraybackslash}X}
\toprule
\textbf{Inputs} & \textbf{Regions} & \textbf{Networks} \\
\midrule
\multirow{5}{*}{Single-feature} 
& SMA.R & Frontal \\
& SFGmed.L & Prefrontal \\
& PCUN.R & Parietal \\
& PUT.L & Subcortical \\
& PUT.R & Subcortical \\
\midrule
\multirow{5}{*}{Multi-feature} 
& PreCG.R & Frontal \\
& SFGdor.R & Prefrontal \\
& SFGdor.L & Prefrontal \\
& SFGmed.L & Prefrontal \\
& PUT.R & Subcortical \\
\bottomrule
\end{tabularx}
\end{minipage}
\hfill
\begin{minipage}[t]{0.48\textwidth}
\centering
\textbf{(B) Ablation study on MRI modalities}
\vspace{1mm}

\begin{tabularx}{\linewidth}{
>{\raggedright\arraybackslash}X
>{\centering\arraybackslash}X
>{\centering\arraybackslash}X
>{\centering\arraybackslash}X
>{\centering\arraybackslash}X}
\toprule
\textbf{Input Modality} & \textbf{Acc} & \textbf{Rec} & \textbf{Pre} & \textbf{F1} \\
\midrule
DFC   & 0.580 & \textbf{1.000} & 0.610 & 0.740 \\
SFC   & 0.690 & 0.930 & 0.680 & 0.760 \\
ALFF  & 0.660 & 0.850 & 0.700 & 0.730 \\
FA    & 0.650 & 0.740 & 0.660 & 0.600 \\
\midrule
\textbf{Dual-MRI} & \textbf{0.825} & 0.900 & \textbf{0.780} & \textbf{0.836} \\
\bottomrule
\end{tabularx}
\end{minipage}

\vspace{1mm}
\parbox{\linewidth}{
\scriptsize
\textit{Note:} 
D-MRI denotes the fused multimodal input using DFC, SFC, ALFF, and FA. 
The FA attribution patterns are model-derived SAM summaries and should not be interpreted as validated clinical biomarkers.
}
\end{table*}

\subsubsection{Regional Attribution Comparison of Grad-CAM, Score-CAM, and SAM}
\label{sec:regional_attribution_comparison}

To provide a more explicit comparison with existing attribution methods, we further summarized regional attribution scores from single-modal Grad-CAM, single-modal Score-CAM, and the proposed multimodal SAM. Grad-CAM and Score-CAM were applied to four separately trained single-feature models using DFC, SFC, ALFF, or FA, respectively. In contrast, SAM was applied to the full multimodal \textbf{\textit{NeuroAlign}} model and generated modality-specific attribution maps for DFC, SFC, ALFF, and FA within the same trained fusion framework. Therefore, this comparison does not aim to claim that SAM is universally superior to existing CAM methods, but to examine whether SAM can provide modality-specific attribution patterns under multimodal fusion while maintaining consistency with single-feature attribution results.

\begin{table*}[htbp]
\centering
\scriptsize
\caption{Comparison of single-modal Grad-CAM, single-modal Score-CAM, and multimodal SAM attribution scores across brain regions. Grad-CAM and Score-CAM scores are obtained from single-feature models, while SAM scores are computed from modality-specific attribution maps of the full multimodal \textbf{\textit{NeuroAlign}} model.}
\label{tab:gradcam_scorecam_sam_region_scores}
\setlength{\tabcolsep}{2.2pt}
\renewcommand{\arraystretch}{1.12}
\begin{tabularx}{\textwidth}{
>{\raggedright\arraybackslash}X
*{12}{>{\centering\arraybackslash}p{0.55cm}}
}
\toprule
\multirow{2}{*}{\textbf{Region / Network}} 
& \multicolumn{4}{c}{\textbf{Grad-CAM}} 
& \multicolumn{4}{c}{\textbf{Score-CAM}} 
& \multicolumn{4}{c}{\textbf{SAM}} \\
\cmidrule(lr){2-5} 
\cmidrule(lr){6-9}
\cmidrule(lr){10-13}
& \textbf{DFC} & \textbf{SFC} & \textbf{ALFF} & \textbf{FA} 
& \textbf{DFC} & \textbf{SFC} & \textbf{ALFF} & \textbf{FA} 
& \textbf{DFC} & \textbf{SFC} & \textbf{ALFF} & \textbf{FA} \\
\midrule
Precuneus                    & 0.78 & 0.62 & 0.71 & --   & 0.80 & 0.65 & 0.74 & --   & \textbf{0.86} & 0.68 & 0.78 & -- \\
Posterior cingulate           & 0.80 & 0.58 & --   & --   & 0.82 & 0.61 & --   & --   & \textbf{0.84} & 0.63 & --   & -- \\
Angular gyrus                 & 0.70 & 0.77 & --   & --   & 0.73 & 0.80 & --   & --   & 0.76 & \textbf{0.84} & --   & -- \\
Inferior parietal lobule      & 0.68 & 0.74 & --   & --   & 0.71 & 0.76 & --   & --   & 0.74 & \textbf{0.80} & --   & -- \\
Parietal cortex               & 0.66 & 0.70 & --   & --   & 0.69 & 0.73 & --   & --   & 0.72 & \textbf{0.77} & --   & -- \\
Temporo-parietal junction     & --   & 0.73 & --   & --   & --   & 0.75 & --   & --   & --   & \textbf{0.78} & --   & -- \\
dlPFC                         & 0.75 & --   & --   & --   & 0.78 & --   & --   & --   & \textbf{0.82} & --   & --   & -- \\
vlPFC                         & 0.72 & --   & --   & --   & 0.75 & --   & --   & --   & \textbf{0.79} & --   & --   & -- \\
Temporal cortex               & 0.64 & 0.66 & --   & --   & 0.66 & 0.68 & --   & --   & 0.69 & 0.71 & --   & -- \\
Occipital cortex              & 0.56 & --   & --   & --   & 0.58 & --   & --   & --   & 0.61 & --   & --   & -- \\
Inferior cerebellum           & 0.82 & 0.84 & --   & --   & 0.84 & 0.86 & --   & --   & 0.87 & \textbf{0.89} & --   & -- \\
Medial cerebellum             & --   & 0.79 & --   & --   & --   & 0.81 & --   & --   & --   & \textbf{0.83} & --   & -- \\
Anterior insula               & --   & 0.60 & --   & --   & --   & 0.62 & --   & --   & --   & \textbf{0.65} & --   & -- \\
Posterior insula              & --   & 0.58 & --   & --   & --   & 0.60 & --   & --   & --   & \textbf{0.63} & --   & -- \\
Middle insula                 & --   & --   & 0.63 & --   & --   & --   & 0.65 & --   & --   & --   & \textbf{0.68} & -- \\
Supplementary motor area      & --   & --   & 0.84 & 0.68 & --   & --   & 0.86 & 0.70 & --   & --   & \textbf{0.89} & 0.73 \\
Medial superior frontal gyrus & --   & --   & 0.80 & 0.76 & --   & --   & 0.82 & 0.78 & --   & --   & \textbf{0.85} & 0.81 \\
Dorsal superior frontal gyrus & --   & --   & --   & 0.83 & --   & --   & --   & 0.85 & --   & --   & --   & \textbf{0.87} \\
Precentral gyrus              & --   & --   & --   & 0.75 & --   & --   & --   & 0.77 & --   & --   & --   & \textbf{0.79} \\
Putamen                       & --   & --   & 0.78 & 0.70 & --   & --   & 0.80 & 0.72 & --   & --   & \textbf{0.82} & 0.74 \\
Thalamus                      & --   & --   & 0.68 & --   & --   & --   & 0.70 & --   & --   & --   & \textbf{0.73} & -- \\
Basal ganglia                 & --   & --   & 0.66 & --   & --   & --   & 0.68 & --   & --   & --   & \textbf{0.71} & -- \\
\bottomrule
\end{tabularx}

\vspace{1mm}
\parbox{\linewidth}{
\scriptsize
\textit{Note:}
All scores are normalized to $[0,1]$. 
Grad-CAM and Score-CAM were applied to single-feature models trained with DFC, SFC, ALFF, or FA separately. 
SAM was applied to the full multimodal \textbf{\textit{NeuroAlign}} model, and the four SAM columns correspond to modality-specific attribution maps generated from DFC, SFC, ALFF, and FA within the multimodal model. 
``--'' indicates that the corresponding region was not prominently highlighted in that attribution map. 

}
\end{table*}

As shown in Table~\ref{tab:gradcam_scorecam_sam_region_scores}, Grad-CAM and Score-CAM provide single-feature attribution patterns, whereas SAM provides modality-specific attribution scores within the full multimodal model. Score-CAM generally produces slightly higher attribution scores than Grad-CAM in several highlighted regions, which is consistent with its gradient-free re-forwarding strategy. SAM further preserves the contribution of each feature type under the same multimodal decision process. For example, DFC- and SFC-related SAM scores mainly emphasize connectivity-domain regions such as the precuneus, posterior cingulate, angular gyrus, inferior parietal lobule, and cerebellar regions, whereas ALFF- and FA-related SAM scores highlight region-domain patterns involving the supplementary motor area, superior frontal regions, putamen, thalamus, and precentral gyrus.

Across DFC, SFC, ALFF, and FA, the attribution maps suggest that different feature types emphasize non-identical but partially complementary patterns. Connectivity-domain features mainly reflect network-level interactions, whereas region-domain features mainly highlight localized functional activity or structural diffusion variations. The recurrent appearance of regions such as the precuneus, posterior cingulate, angular gyrus, parietal cortex, frontal regions, cerebellar regions, and putamen-related areas suggests potential cross-feature consistency. However, these findings should be interpreted as model-derived attribution patterns rather than direct clinical evidence or validated causal biomarkers.

\subsection{Ablation study}
\label{sec:5.3}

\subsubsection{Ablation Study on Model Architectures}
\label{sec:5.3.1}

\begin{table*}[htbp]
\scriptsize
    \centering
    % \small
    \setlength{\tabcolsep}{1.8mm}
\caption{Hierarchical ablation study of \textbf{\textit{NeuroAlign}} components. The last row denotes an unaligned four-feature fusion variant using DFC, SFC, ALFF, and FA without DMHA or DDHI, rather than a conventional single-modality baseline.}
    \label{tab: Table4}
       \begin{tabularx}{\textwidth}{*{9}{>{\centering\arraybackslash}X}}
     \toprule
     \multicolumn{3}{c}{\textit{DMHA}} & \multicolumn{2}{c}{\textit{DDHI}} & \multirow{2}{*}{Accuracy} & \multirow{2}{*}{Recall} & \multirow{2}{*}{Precision} & \multirow{2}{*}{F1-score}\\
     \cmidrule(l){1-3} \cmidrule(l){4-5}
     TSA & DSA & FSA & FI & GI \\
     \midrule
     \checkmark & \checkmark & \checkmark & \checkmark & \checkmark 
     & \textbf{0.791} & \textbf{0.867} & \textbf{0.778} & \textbf{0.820}\\
     \midrule
     \checkmark & \checkmark & \checkmark & \checkmark & \ding{55} 
     & 0.750 & 0.850 & 0.765 & 0.805\\
     \midrule
     \checkmark & \checkmark & \checkmark & \ding{55} & \ding{55} 
     & 0.724 & 0.790 & 0.735 & 0.761\\ 
     \midrule
     \checkmark & \checkmark & \ding{55} & \ding{55} & \ding{55} 
     & 0.690 & 0.720 & 0.692 & 0.706\\
     \midrule
     \checkmark & \ding{55} & \ding{55} & \ding{55} & \ding{55} 
     & 0.620 & 0.640 & 0.610 & 0.625\\
     \midrule
     \ding{55} & \ding{55} & \ding{55} & \ding{55} & \ding{55} 
     & 0.565 & 0.470 & 0.590 & 0.523\\
     \bottomrule
    \end{tabularx}
\vspace{1mm}
\parbox{\linewidth}{
\scriptsize
\textit{Note:}
The variant without DMHA and DDHI still uses the four input features, but removes all explicit alignment and interaction modules. 
It is used to evaluate the effect of unstructured multimodal fusion and should not be interpreted as a standard single-modality baseline.
}
\end{table*}

\begin{figure}
  \centering
  \includegraphics[width=0.99\linewidth]{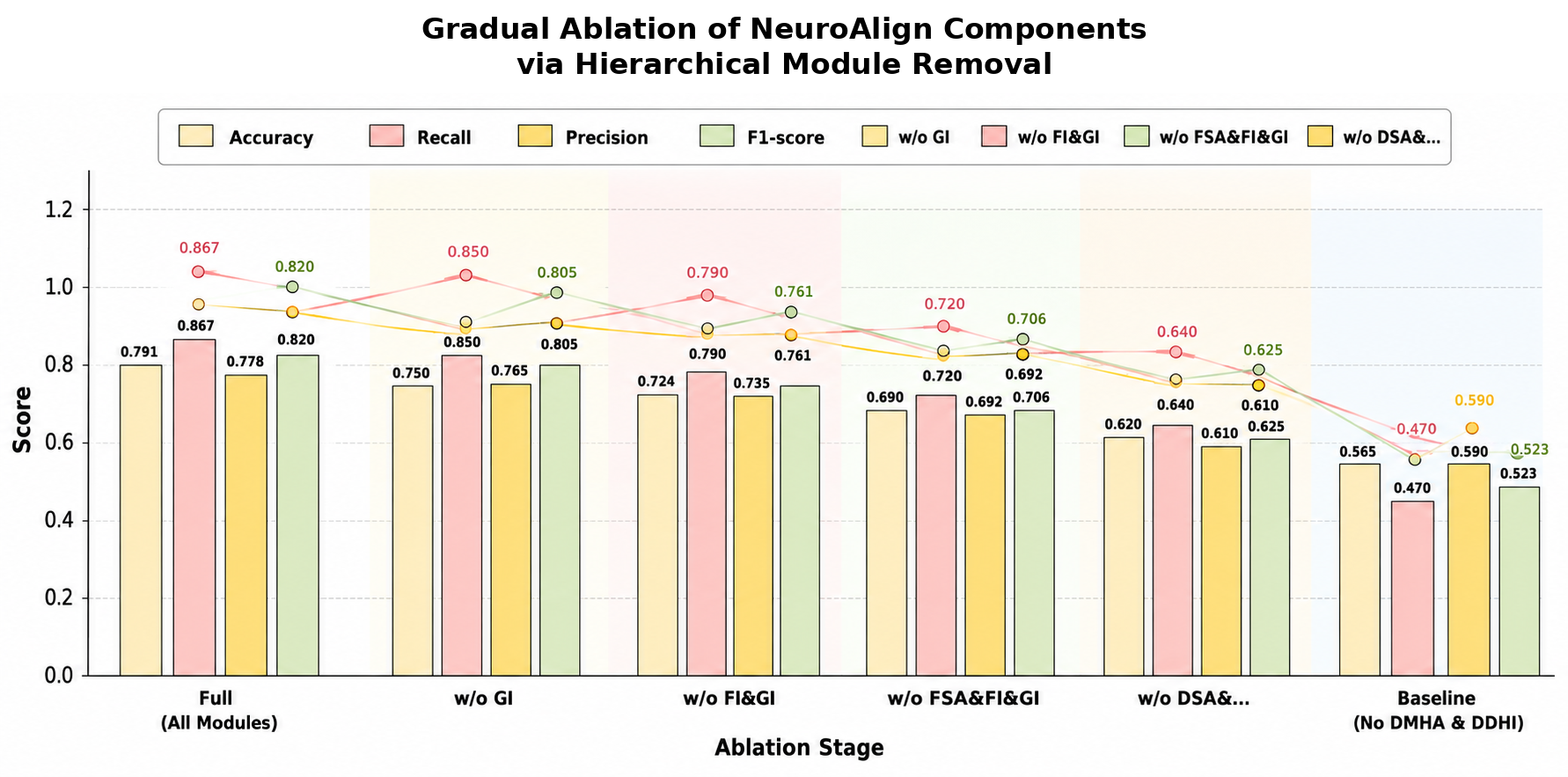}
  \caption{Gradual ablation of \textit{NeuroAlign} components. 
    Each curve shows performance across four metrics; vertical shaded regions indicate the module(s) removed at each stage. 
    The full model (\textit{Full}) achieves highest accuracy and F1-score; baseline shows low recall.}
  \label{fig:ablation_clean}
\end{figure}

% \begin{table}[htbp]
% \caption{Ablation study on MRI modalities: performance comparison of individual and fused inputs.}
% \label{tab:ablation_modality}
% \centering
% \footnotesize
% \setlength{\tabcolsep}{6pt}        % 增加列间距，避免拥挤
% \renewcommand{\arraystretch}{1.2} % 行高微调，提升视觉舒适度
% \begin{tabular}{l cccc}
%     \toprule
%     \textbf{Input Modality} & \textbf{Acc} & \textbf{Rec} & \textbf{Pre} & \textbf{F1} \\
%     \midrule
%     DFC   & 0.580 & \textbf{1.000} & 0.610 & 0.740 \\
%     SFC   & 0.690 & 0.930 & 0.680 & 0.760 \\
%     ALFF  & 0.660 & 0.850 & 0.700 & 0.730 \\
%     FA    & 0.650 & 0.740 & 0.660 & 0.600 \\
%     \midrule
%     \textbf{D-MRI} & \textbf{0.825} & 0.900 & \textbf{0.780} & \textbf{0.836} \\
%     \bottomrule
% \end{tabular}
% \end{table}

\begin{figure}
  \centering
  \includegraphics[width=0.99\linewidth]{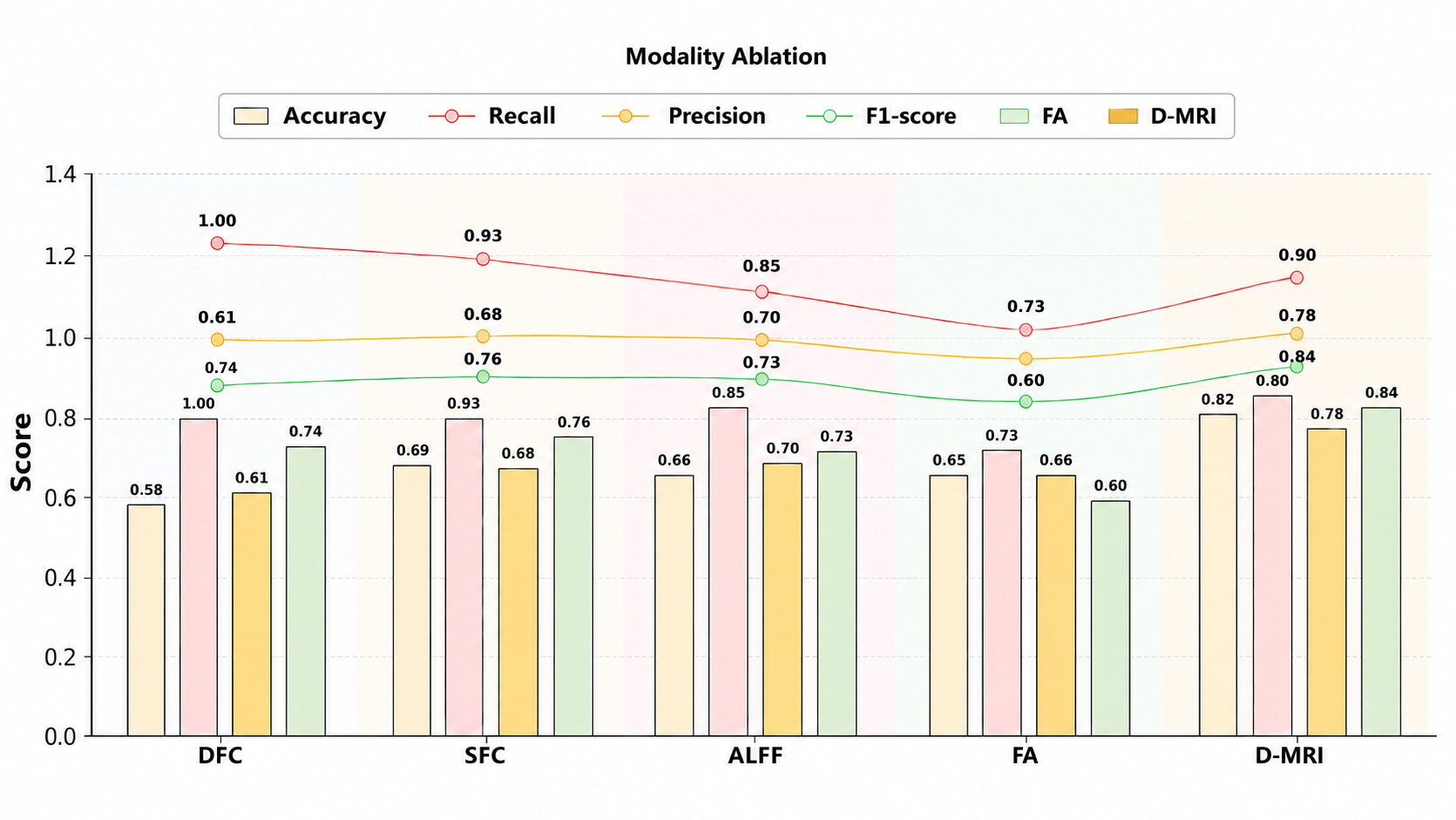}
  \caption{Performance comparison across MRI input modalities. 
Each curve shows Accuracy, Recall, Precision, and F1-score for DFC, SFC, ALFF, FA, and fused D-MRI. 
D-MRI achieves the highest accuracy and F1-score in this experiment, suggesting the benefit of combining complementary features in our NeuroAlign.}
  \label{fig:modality_trend}
\end{figure}

% To validate the necessity of each component of \textbf{\textit{NeuroAlign}}, we evaluated the model's performance on the ADNI dataset by hierarchically removing the alignments (\textit{DMHA: TSA|DSA|FSA}) and interactions (\textit{DDHI: FI|GI}), as detailed in Table \ref{tab: Table4} from top to bottom. The cross-domain attention modules in Fig. \ref{fig: Fig2} were removed to observe the effects of \textit{GI} in \textit{DDHI}. Furthermore, the outputs of \textit{DMHA} bypassed \textit{FI} and were directly fed to the final residual projector to validate the entire \textit{DDHI} component. Subsequently, we methodically removed the contrastive restrictions in \textit{FSA} and \textit{DSA}, as well as the pyramid adaptive convolution pipeline in \textit{TSA} to assess the impacts of the hierarchical alignments within the \textit{DMHA} component.

Table~\ref{tab: Table4} should be interpreted as a hierarchical removal analysis rather than a comparison with conventional single-modality baselines. The last row, where all DMHA and DDHI components are removed, is not a standard single-modality model. Instead, it corresponds to an unaligned four-feature fusion setting in which DFC, SFC, ALFF, and FA are jointly used but without explicit temporal alignment, dynamic-static alignment, functional-structural alignment, or connectivity-region interaction. Therefore, its low accuracy does not indicate that simple single-modality models are necessarily weak; rather, it shows that directly combining heterogeneous multimodal features without structured alignment may introduce feature conflict and severe imbalance between sensitivity and specificity.

This interpretation is supported by the metric distribution of the no-alignment/no-interaction variant, which obtains high precision but very low recall. Such behavior suggests that the model becomes conservative and fails to detect many positive cases when heterogeneous features are fused without proper alignment. In contrast, adding TSA, DSA, FSA, FI, and GI progressively improves both accuracy and F1-score, indicating that the proposed modules mainly help transform the four heterogeneous inputs into more compatible and discriminative representations. Thus, the ablation study is intended to demonstrate the necessity of structured multimodal fusion, not to claim that the unstructured four-feature variant is a competitive baseline.

\subsubsection{Ablation Study on Feature Modalities}
\label{sec:appendix abaltion}
 Table~\ref{tab:fa_attribution_modality_ablation}B summarizes the contribution of each component under a hierarchical removal setting. Intriguingly, the peak in recall was achieved when the final connectivity-regional interaction was omitted, while the lowest recall accompanied by the highest precision was noted when employing DFC, SFC, ALFF, and FA as inputs, but without any alignments or interactions. 
These phenomena suggest:
\begin{enumerate}[label=$\bullet$, itemsep=0mm, parsep=0mm, topsep=0mm, partopsep=0mm]
    \item Various features from dual-modal MRI can challenge the generalizability of modeling, potentially leading to numerous missed detections. The hierarchical alignment modules appear to reduce these difficulties; 
    \item Attention-based cross-domain interactions further improve the balance between sensitivity and specificity in this setting.
\end{enumerate}

\subsubsection{Ablation Study on Alignment Objectives}
\label{sec:alignment_objective_ablation}

To further examine whether the proposed contrastive alignment objectives are necessary for DSA and FSA, we compared them with two simpler alignment alternatives under the same backbone, multimodal input setting, and training protocol. Specifically, cosine similarity loss and maximum mean discrepancy (MMD) loss were used as representative alternatives to the contrastive objective. Cosine similarity loss serves as an instance-level positive-pair alignment objective, which only encourages paired representations to be close. MMD loss serves as a distribution-level alignment objective, which matches the global distributions of two representation spaces. These two alternatives were selected to evaluate whether the improvement of \textbf{\textit{NeuroAlign}} is mainly due to generic feature alignment or to the discriminative subject-level alignment introduced by contrastive learning.

For cosine alignment, the objective is defined as:
\begin{equation}
    \mathcal{L}_{\mathrm{cos}}(z_a,z_b)
    =
    1 -
    \frac{z_a^\top z_b}{\|z_a\|\|z_b\|},
\end{equation}
where $z_a$ and $z_b$ denote paired representations from the same subject. 
In the DSA comparison, they correspond to DFC and SFC embeddings. 
In the FSA comparison, they correspond to the two functional--structural fusion probes.

For MMD alignment, the objective is defined as:
\begin{equation}
\begin{aligned}
    \mathcal{L}_{\mathrm{MMD}}
    =
    &\frac{1}{n^2}\sum_{i,j} k(z_a^i,z_a^j)
    + \frac{1}{n^2}\sum_{i,j} k(z_b^i,z_b^j) \\
    &- \frac{2}{n^2}\sum_{i,j} k(z_a^i,z_b^j),
\end{aligned}
\end{equation}
where $k(\cdot,\cdot)$ denotes the RBF kernel. 
Unlike cosine similarity loss, MMD performs distribution-level matching. 
Unlike contrastive learning, however, it does not explicitly preserve subject-level positive-pair correspondence or separate mismatched samples.

As shown in Table~\ref{tab:alignment_objective_ablation}(a), when only the DSA alignment objective is changed and the other modules are kept unchanged, both cosine similarity loss and MMD loss improve over removing the DSA alignment loss. This indicates that aligning dynamic and static functional representations is beneficial for multimodal fusion. However, the contrastive objective achieves the best overall performance, suggesting that subject-level dynamic--static alignment with both positive-pair attraction and negative-pair separation is more effective than simple similarity matching or global distribution alignment.

Table~\ref{tab:alignment_objective_ablation}(b) further compares different alignment objectives in FSA. Similarly, replacing the FSA contrastive objective with cosine similarity or MMD improves performance compared with removing the FSA alignment loss, indicating that functional--structural alignment is important. Nevertheless, contrastive FSA achieves the highest accuracy and F1-score. These results suggest that the element-wise product and summation probes benefit from a discriminative contrastive constraint, rather than merely from being regularized by a generic similarity or distribution-matching objective.

As shown in Table~\ref{tab:alignment_objective_ablation}, removing the alignment loss leads to the lowest overall performance, suggesting that directly combining heterogeneous DFC, SFC, ALFF, and FA features without explicit alignment may introduce representation inconsistency. Replacing the contrastive objective with cosine similarity loss improves the performance, indicating that instance-level positive-pair alignment is helpful for multimodal fusion. However, cosine alignment only encourages paired features from the same subject to be close and does not explicitly separate mismatched subjects or modalities, which may limit its discriminative ability.

\begin{table*}[htbp]
\centering
\small
\caption{Ablation study of alignment objectives in DSA and FSA under the same backbone and multimodal input setting.}
\label{tab:alignment_objective_ablation}
\setlength{\tabcolsep}{5pt}
\renewcommand{\arraystretch}{1.15}

\begin{minipage}{0.48\textwidth}
\centering
\textbf{(a) Alignment objectives in DSA}
\vspace{1mm}

\begin{tabular}{lcccc}
\toprule
\textbf{DSA Objective} & \textbf{Acc} & \textbf{Rec} & \textbf{Pre} & \textbf{F1} \\
\midrule
w/o DSA alignment loss & 0.724 & 0.724 & 0.690 & 0.707 \\
Cosine similarity loss & 0.775 & 0.800 & 0.762 & 0.781 \\
MMD loss               & 0.748 & 0.760 & 0.735 & 0.747 \\
Contrastive loss       & \textbf{0.825} & \textbf{0.900} & \textbf{0.780} & \textbf{0.836} \\
\bottomrule
\end{tabular}
\end{minipage}
\hfill
\begin{minipage}{0.48\textwidth}
\centering
\textbf{(b) Alignment objectives in FSA}
\vspace{1mm}

\begin{tabular}{lcccc}
\toprule
\textbf{FSA Objective} & \textbf{Acc} & \textbf{Rec} & \textbf{Pre} & \textbf{F1} \\
\midrule
w/o FSA alignment loss & 0.750 & 0.800 & 0.727 & 0.762 \\
Cosine similarity loss & 0.802 & 0.860 & 0.767 & 0.811 \\
MMD loss               & 0.775 & 0.820 & 0.745 & 0.781 \\
Contrastive loss       & \textbf{0.825} & \textbf{0.900} & \textbf{0.780} & \textbf{0.836} \\
\bottomrule
\end{tabular}
\end{minipage}

\vspace{1mm}
\parbox{\linewidth}{
\scriptsize
\textit{Note:}
All variants use the same DFC+SFC+ALFF+FA inputs and the same network backbone. 
In subtable (a), only the alignment objective in DSA is changed, while TSA, FSA, FI, and GI are kept unchanged. 
In subtable (b), only the alignment objective in FSA is changed, while TSA, DSA, FI, and GI are kept unchanged. 
Cosine similarity loss performs positive-pair feature alignment, MMD loss performs distribution-level alignment, and contrastive loss performs subject-level alignment with both positive-pair attraction and negative-pair separation. 
Results are reported as average performance under the five-fold cross-validation protocol.
}
\end{table*}

MMD loss also improves over the no-alignment variant, suggesting that distribution-level alignment can reduce the discrepancy between feature spaces. Nevertheless, its performance is lower than that of cosine similarity loss and contrastive loss. This may be because MMD focuses on matching global distributions but does not explicitly preserve subject-level correspondence, which is important for cognitive impairment analysis where individual-specific functional--structural patterns may carry discriminative information.

Compared with these two simpler alternatives, the proposed contrastive alignment achieves the best overall performance. This result suggests that the contrastive objective is more suitable for DSA and FSA because it simultaneously pulls positive pairs closer and pushes mismatched pairs apart. Therefore, the improvement of \textbf{\textit{NeuroAlign}} is not merely caused by imposing a generic similarity constraint, but is more closely related to subject-level discriminative alignment between dynamic--static and functional--structural representations.

\subsection{Cross-dataset Validation}
\label{sec:5.4}

\begin{table*}[htbp]
\caption{Cross-dataset validation of \textit{NeuroAlign} among GUTCM, ADNI, and OASIS under five-fold cross-validation. Models trained in each fold were transferred to external datasets for evaluation, and the reported results are averaged across five folds.}
\label{tab:cross_dataset}
\centering
\small
\setlength{\tabcolsep}{4pt}
\renewcommand{\arraystretch}{1.2}
\begin{tabularx}{\textwidth}{*{10}{>{\centering\arraybackslash}X}}
    \toprule
    \multicolumn{2}{c}{\textbf{GUTCM}} &
    \multicolumn{2}{c}{\textbf{ADNI}} &
    \multicolumn{2}{c}{\textbf{OASIS}} &
    \multirow{2}{*}{\textbf{ACC}} &
    \multirow{2}{*}{\textbf{PRE}} &
    \multirow{2}{*}{\textbf{REC}} &
    \multirow{2}{*}{\textbf{F1}} \\
    \cmidrule(lr){1-2}
    \cmidrule(lr){3-4}
    \cmidrule(lr){5-6}
    \textbf{Train} & \textbf{Test} &
    \textbf{Train} & \textbf{Test} &
    \textbf{Train} & \textbf{Test} &
    & & & \\
    \midrule
    \checkmark & --         & --         & \checkmark & --         & --         & 0.650 & 0.650 & 0.650 & 0.650 \\
    \checkmark & --         & --         & --         & --         & \checkmark & 0.704 & 0.842 & 0.958 & 0.896 \\
    --         & \checkmark & \checkmark & --         & --         & --         & 0.656 & 0.562 & 0.729 & 0.624 \\
    --         & --         & \checkmark & --         & --         & \checkmark & 0.696 & 0.835 & 0.958 & 0.892 \\
    --         & \checkmark & --         & --         & \checkmark & --         & 0.584 & 0.527 & 0.615 & 0.568 \\
    --         & --         & --         & \checkmark & \checkmark & --         & 0.603 & 0.592 & 0.625 & 0.608 \\
    \checkmark & --         & \checkmark & --         & --         & \checkmark & 0.712 & 0.858 & 0.958 & 0.905 \\
    --         & \checkmark & \checkmark & --         & \checkmark & --         & 0.642 & 0.571 & 0.708 & 0.632 \\
    \bottomrule
\end{tabularx}

\vspace{1mm}
\parbox{\linewidth}{
\scriptsize
\textit{Note:}
This transfer experiment was performed under the five-fold cross-validation protocol. 
For each fold, the model was trained on the dataset(s) marked as ``Train'' and directly evaluated on the external dataset marked as ``Test'' without using target-domain labels during training. 
The reported values denote the average transfer performance across the five fold-specific models. 
Although the OASIS subset was expanded to 90 subjects after applying the paired multimodal inclusion criteria, results involving OASIS should still be interpreted cautiously because its scale remains smaller than GUTCM and ADNI.
}
\end{table*}

In addition to the comparative studies conducted on public datasets, we further perform \textbf{cross-dataset validation experiments} to evaluate the preliminary transferability of our model across different clinical sites. As shown in Table~\ref{tab:cross_dataset}, transfer performance varies across source-target settings. Models trained on GUTCM or ADNI show comparable transfer performance when OASIS is used as the target dataset, whereas models trained only on OASIS obtain relatively lower transfer performance on GUTCM and ADNI. This pattern is consistent with the smaller scale of the OASIS training subset after paired multimodal inclusion criteria. Overall, the transfer accuracy remains lower than the corresponding within-dataset performance in most settings, suggesting that substantial site-related distribution shifts still exist. Therefore, these results should be interpreted as preliminary cross-site evaluation rather than evidence of reliable real-world multi-center deployment.

\subsection{Domain Gap Analysis for Cross-dataset Validation}
\label{sec:domain_gap_analysis}

The cross-dataset validation results show that transfer performance varies across different source-target settings and remains lower than the corresponding within-dataset performance in most cases. 
Therefore, we do not interpret these results as evidence that \textbf{\textit{NeuroAlign}} fully solves multi-center domain generalization. 
Instead, the results suggest that the model retains partial transferability across heterogeneous datasets, while non-negligible domain shifts still exist.

The performance degradation is likely associated with differences in acquisition protocols across datasets. 
As summarized in Table~\ref{tab:infor}, GUTCM, ADNI, and OASIS differ in TE, FOV, slice thickness, flip angle, number of slices, and DTI acquisition settings. 
These protocol differences may influence fMRI-derived connectivity features, including DFC and SFC, regional functional activity features such as ALFF, and diffusion-derived structural features such as FA. 
Consequently, the cross-dataset performance drop may reflect scanner- and protocol-induced feature distribution shifts, rather than only model overfitting to a single dataset.

To further inspect the feature-level distribution discrepancy, we visualized dataset-wise distributions using feature-specific t-SNE projections, as shown in Fig.~\ref{fig:domain_gap_tsne}. 
Because DFC, SFC, ALFF, and FA have different dimensionalities and feature semantics, each feature type was projected into an independent t-SNE space instead of being mixed into a single embedding space. 
As observed in the four subplots, the distributions of GUTCM, ADNI, and OASIS are not completely separated; partial overlap exists in each feature space, indicating that the datasets still share common feature patterns. 
However, dataset-dependent shifts are also visible, especially in the relative positions of ADNI and OASIS compared with GUTCM across DFC, SFC, ALFF, and FA. 
The FA subplot shows a relatively clearer separation among datasets, which may be related to the larger differences in DTI acquisition parameters. 
These observations are consistent with the moderate cross-dataset transfer performance and suggest that the degradation mainly arises from residual multi-site domain gaps rather than a complete failure of feature alignment.

\begin{figure}
  \centering
  \includegraphics[width=0.85\linewidth]{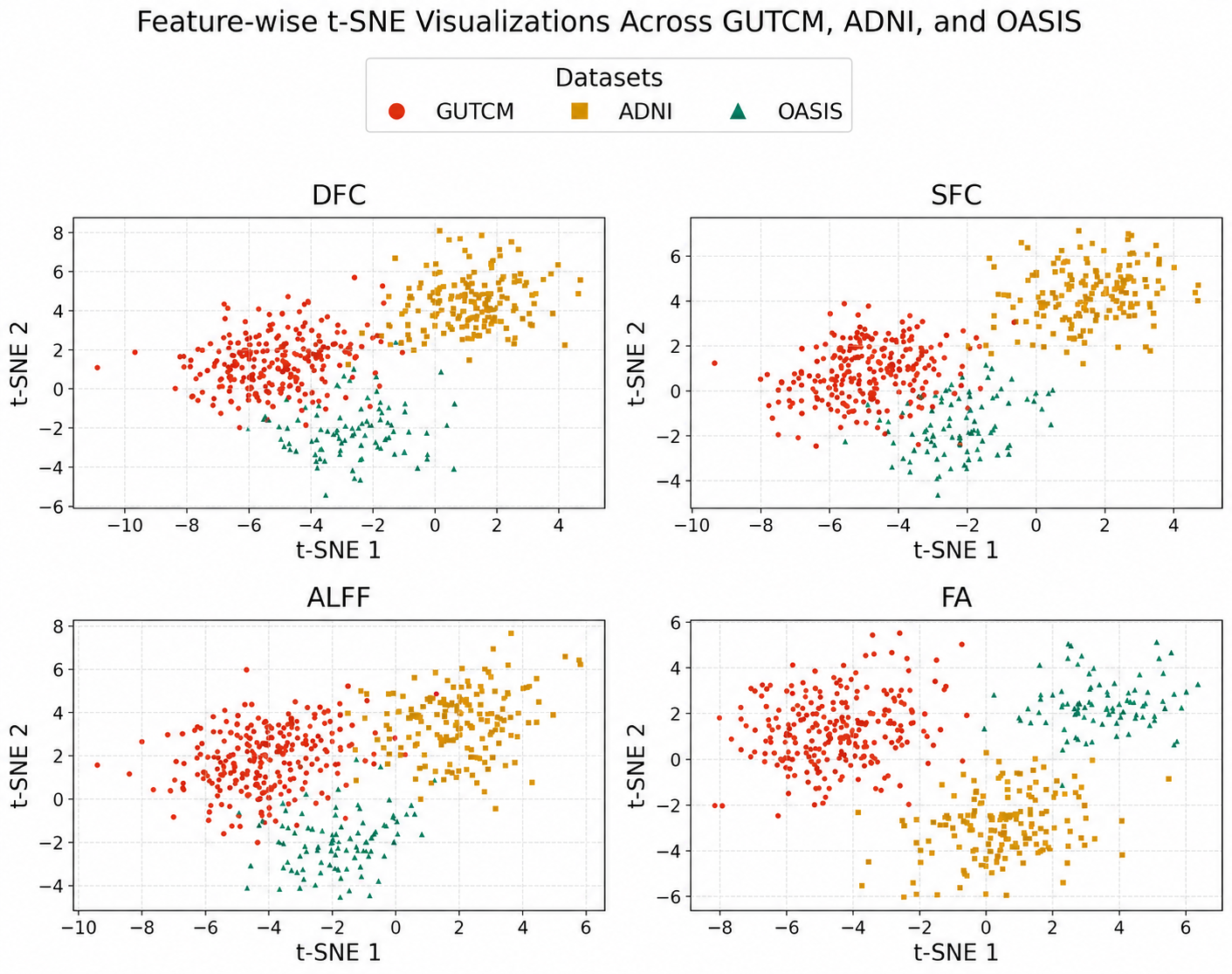}
  \caption{Feature-wise t-SNE visualization of dataset distributions across GUTCM, ADNI, and OASIS. 
  Each subplot corresponds to an independent feature space, including DFC, SFC, ALFF, and FA. 
  Different colors and markers denote different datasets. 
  The partial overlap indicates shared feature patterns across datasets, whereas the visible dataset-dependent shifts suggest residual domain gaps caused by heterogeneous acquisition protocols.}
  \label{fig:domain_gap_tsne}
\end{figure}

\section{Discussion and Conclusion}
\label{sec:discussion}

We present \textbf{\textit{NeuroAlign}}, a structured multimodal fusion framework for cognitive impairment detection that leverages hierarchical alignment (DMHA) and dynamic interaction (DDHI) across multi-modal MRI features spanning structural-functional, static-dynamic, and temporal domains. Evaluations on GUTCM, ADNI, and OASIS suggest that \textit{NeuroAlign} achieves competitive performance under modality-controlled and dataset-specific reference comparisons. The proposed Synergistic Activation Mapping (SAM) provides feature-level attribution maps for DFC, SFC, ALFF, and FA, highlighting model-relevant patterns involving regions and networks frequently discussed in cognitive impairment studies, including DMN/FPN-related regions. These attribution maps should be interpreted as model-derived associations rather than clinically validated causal biomarkers.

Despite these promising results, several limitations remain. First, fMRI/DTI based diagnosis continues to face well-documented challenges: (1) scarcity of labeled cognitive impairment cases in publicly available datasets; (2) elevated risk of overfitting due to limited sample sizes; and (3) computationally intensive preprocessing pipelines that constrain scalability and real-time deployment. Although our study aggregates data from three datasets with over 500 participants — a relatively large cohort in neuroimaging standards — this scale remains insufficient to fully ensure model generalizability across global populations. Achieving true population-level robustness would require datasets on the order of tens of thousands of subjects, which are currently infeasible to acquire, annotate, and preprocess with existing infrastructure.

A specific limitation concerns the OASIS dataset. Although the OASIS subset was expanded to 90 subjects after applying the paired multimodal inclusion criteria, it remains smaller than GUTCM and ADNI. This may still limit the statistical power of OASIS-based evaluation and cross-dataset transfer analysis. Therefore, the OASIS results should be interpreted as preliminary external validation rather than definitive evidence of large-scale generalizability. Future work should prioritize the integration of additional fMRI and DTI biomarkers from OASIS, particularly longitudinal scans and harmonized preprocessing outputs, to strengthen validation on this benchmark dataset and further test the transportability of \textit{NeuroAlign} across heterogeneous acquisition protocols.

It is important to note that the performance gain of \textbf{\textit{NeuroAlign}} on ADNI is modest when compared with the strongest previously reported method. Therefore, the contribution of the proposed framework should not be understood solely as a large numerical improvement on a single benchmark. Instead, our results suggest that hierarchical multimodal alignment is beneficial in three complementary aspects. First, it provides more balanced classification behavior, particularly by improving recall while maintaining competitive F1-score. Second, it shows consistent performance across ADNI, OASIS, and GUTCM, where acquisition protocols and cohort characteristics differ. Third, the ablation results indicate that the proposed alignment and interaction modules alleviate the difficulty of directly fusing heterogeneous DFC, SFC, ALFF, and FA features.

The relatively low performance of the variant without DMHA and DDHI also deserves careful interpretation. This variant should not be regarded as a simple single-modality baseline. Instead, it represents a naive four-feature fusion setting, where dynamic connectivity, static connectivity, regional functional activity, and structural diffusion features are combined without explicit alignment or cross-domain interaction. Since these features differ substantially in dimensionality, statistical distribution, biological meaning, and noise characteristics, direct fusion can increase optimization difficulty and may cause the classifier to rely on incomplete or biased evidence. This is reflected by the high precision but low recall of the no-alignment/no-interaction variant, indicating that many positive MCI cases are missed. The gradual recovery of accuracy and F1-score after introducing FSA, DSA, TSA, FI, and GI suggests that the proposed modules are useful for reducing feature incompatibility and improving multimodal representation learning.

Nevertheless, we acknowledge that the current results do not imply that the proposed hierarchical design is universally superior to all simpler alternatives. In particular, the accuracy improvement on ADNI over the best reported external baseline is small, and external SOTA comparisons may involve differences in preprocessing, input features, and evaluation protocols. Future work will further validate the proposed framework using larger cohorts, repeated cross-validation with confidence intervals, and additional modality-controlled baselines such as simple concatenation, late fusion, and lightweight attention-based fusion. These analyses will help clarify the trade-off between model complexity and performance gain.

The cross-dataset results should be interpreted cautiously. Although \textbf{\textit{NeuroAlign}} shows partial transferability among GUTCM, ADNI, and OASIS, the transfer performance varies across source-target settings and remains below within-dataset performance in most cases. In particular, OASIS-related transfer results should be interpreted with caution because the expanded OASIS subset remains smaller than GUTCM and ADNI. These results do not yet demonstrate reliable deployment in real-world multi-center clinical settings. Instead, they suggest that structured functional--structural alignment can partially improve transferability under heterogeneous acquisition protocols, while scanner- and protocol-related shifts remain a major challenge. Future studies should evaluate the model on larger multi-center cohorts, include harmonization strategies such as ComBat or domain-adversarial learning, and conduct prospective validation before clinical translation.

In conclusion, \textit{NeuroAlign} provides a structured approach for multimodal neuroimaging fusion by modeling cross-domain alignment and interaction among DFC, SFC, ALFF, and FA features. The current results suggest potential benefits for multimodal representation learning and feature-level attribution, but they do not establish clinical utility or population-level generalizability. Future work will require larger multi-center cohorts, harmonization strategies, prospective validation, and quantitative interpretability benchmarks before the framework can be considered for clinical translation.

% \section{CRediT authorship contribution statement}
% \textbf{Xiongri Shen}: theoretical development, data analysis, drafting the article.
% \textbf{Jiaqi Wang}: theoretical development, revising the article. 
% % \textbf{Linling Li}: experimental design, drafting the article.
% \textbf{Zhenxi Song}: theoretical development, interpretation of data, revising the article. 
% \textbf{Leilei Zhao}: theoretical development, revising the article. 
% \textbf{Yi Zhong}: experimental design, data collection and pre-processing.
% \textbf{Yichen Wei}: experimental design, data collection and pre-processing. 
% \textbf{Lingyan Liang}: experimental design, data collection and pre-processing. 
% % \textbf{Chenqi Xu}: experimental design, data collection and pre-processing. 
% % \textbf{MCADI}: experimental design, data collection and pre-processing.
% \textbf{Baiying Lei}: experimental design, interpretation of data drafting the article. 
% \textbf{Shuqiang Wang}: experimental design, interpretation of data drafting the article. 
% \textbf{Demao Deng}: experimental design, interpretation of data drafting the article. 
% \textbf{Zhiguo Zhang}: theoretical development, experimental design, interpretation of data, revising the article. All authors have approved the final version of the article.

\section{Acknowledgment}
This research was supported by the the National Natural Science Foundation of China (Grants 62306089, 32361143787, 82102032), the key Project of Basic Research of Shenzhen (NO: JCYJ20200109113603854), the Shenzhen Science and Technology Program (Grant No. RCBS20231211090800003), the Shenzhen Science and Technology Program (ZDSYS20230626091203008), Shenzhen-Hong Kong Institute of Brain Science-Shenzhen Fundamental Research Institutions (2023SHIBS0003), and the Guangxi Natural Science Foundation (Grant No. 2023GXNS-FBA026073).

\section{Declaration of competing interest}
The authors declare that they have no known competing financial interests or personal relationships that could have appeared to 
influence the work reported in this paper.

% Use the style of numbering in square brackets.
% If nothing is used, default style will be taken.
%\begin{enumerate}[a)]
%\item 
%\item 
%\item 
%\end{enumerate}  

%\begin{itemize}
%\item 
%\item 
%\item 
%\end{itemize}  

%\begin{description}
%\item[]
%\item[] 
%\item[] 
%\end{description}  

% \clearpage %%Remove this from your manuscript

% Figure
% \begin{figure}%[]
%   \centering
% %    \includegraphics{}
%     \caption{}\label{fig1}
% \end{figure}

% \begin{table}%[]
% \begin{tabular*}{\tblwidth}{@{}LL@{}}
% \toprule
% \midrule
%  & \\
%  & \\
%  & \\
%  & \\
% \bottomrule
% \end{tabular*}
% \end{table}

% Uncomment and use as the case may be
%\begin{theorem} 
%\end{theorem}

% Uncomment and use as the case may be
%\begin{lemma} 
%\end{lemma}


\section{CRediT authorship contribution statement}\label{}

\printcredits
\textbf{Xiongri Shen}: theoretical development, data analysis, drafting the article.
\textbf{Zhenxi Song}: theoretical development, interpretation of data, revising the article. 
\textbf{Linling Li}: experimental design, drafting the article.
\textbf{Jiaqi Wang}: theoretical development, revising the article. 
\textbf{Leilei Zhao}: theoretical development, revising the article. 
\textbf{Yi Zhong}: experimental design, data collection and pre-processing.
\textbf{Yichen Wei}: experimental design, data collection and pre-processing. 
\textbf{Lingyan Liang}: experimental design, data collection and pre-processing. 
\textbf{Chenqi Xu}: experimental design, data collection and pre-processing. 
\textbf{Baiying Lei}: experimental design, interpretation of data drafting the article. 
\textbf{Shuqiang Wang}: experimental design, interpretation of data drafting the article. 
\textbf{Demao Deng}: experimental design, interpretation of data drafting the article. 
\textbf{Zhiguo Zhang}: theoretical development, experimental design, interpretation of data, revising the article. All authors have approved the final version of the article.
\bibliographystyle{cas-model2-names}

\bibliography{cas-refs}



\end{document}